\documentclass[10pt,twocolumn,letterpaper]{article}

\usepackage[pagenumbers]{cvpr} 
\usepackage{multirow} 
\usepackage[dvipsnames]{xcolor}

\definecolor{cvprblue}{rgb}{0.21,0.49,0.74}
\usepackage[pagebackref,breaklinks,colorlinks,citecolor=cvprblue]{hyperref}

\def\paperID{****} 
\def\confName{CVPR}
\def\confYear{2024}

\title{All in One: RGB, RGB-D, and RGB-T Salient Object Detection}

\author{Xingzhao Jia\\
Hefei University of Technology\\
Hefei, China\\
{\tt\small jxz625@126.com}
\and
Zhongqiu Zhao*\\
Hefei University of Technology\\
Hefei, China\\
{\tt\small z.zhao@hfut.edu.cn}
\and
Changlei Dongye\\
Shandong University of Science and Technology\\
Qingdao, China\\
{\tt\small dycl@sdust.edu.cn}
\and
Zhao Zhang\\
Hefei University of Technology\\
Hefei, China\\
{\tt\small cszzhang@gmail.com}
}

\begin{document}
\maketitle
\begin{abstract}
Salient object detection (SOD) aims to identify the most attractive objects within an image. Depending on the type of data being detected, SOD can be categorized into various  forms, including RGB, RGB-D (Depth), RGB-T (Thermal) and light field SOD. Previous researches have focused on saliency detection with individual data type.  If the RGB-D SOD model is forced to detect RGB-T data it will perform poorly. We  propose an innovative model framework that provides a unified solution for the salient object detection task of three types of data (RGB, RGB-D, and RGB-T). The three types of data can be handled in one model (all in one) with the same weight parameters. In this framework, the three types of data are concatenated in an ordered manner within a single input batch, and features are extracted using a transformer network. Based on this framework, we propose an efficient lightweight SOD model, namely AiOSOD, which can detect any RGB, RGB-D, and RGB-T data with high speed (780FPS for RGB data, 485FPS for RGB-D or RGB-T data). Notably, with only 6.25M parameters, AiOSOD achieves excellent  performance on RGB, RGB-D, and RGB-T datasets.  
\end{abstract}    
\section{Introduction}
\label{sec:intro}

Salient Object Detection (SOD) can locate the most salient objects in an image, and it is widely used as an important preprocessing method for many vision tasks such as image/video segmentation \cite{segmentation1, segmentation2}, video compression \cite{compression}, and visual tracking \cite{ retrieval}. Depending on the data to be processed, SOD can be divided into 2D (RGB) SOD and 3D (RGB-D, RGB-T) SOD. 3D SOD addresses challenging scenarios by introducing depth maps or thermal maps paired with RGB maps. 

Given the detection requirements for RGB, RGB-D, and RGB-T data, current models \cite{ICON,ITSD,liu2021vst,fu2021jldcf,liu2021swinnet,zhou2023lsnet, DIGR-Net} can only handle a single type of data, and it cannot handle these three types of data simultaneously.
Therefore, it is necessary to develop a model framework that can simultaneously meet the detection requirements for these three types of data. The model adopting this framework only needs to be trained once, and then it can use the same set of weight parameters to detect RGB, RGB-D, and RGB-T data.

In general, RGB SOD \cite{RCSB,ICON,ITSD} requires only one network to extract RGB features, while RGB-D or RGB-T SOD \cite{liu2021swinnet,liu2021vst,zhou2023lsnet} requires two networks to extract features from  two modalities separately. The 3D SOD model differs from the 2D SOD model in that it includes an additional feature learning network and a multi-modal feature fusion module. To achieve the detection of three types of data in one model, this model needs to be optimized based on a 3D SOD framework to effectively handle RGB, RGB-D, and RGB-T data.
When processing RGB data, the multi-modal feature fusion module has less impact on the RGB saliency prediction results because it processes learned RGB features, which is equivalent to fusing multiple RGB features. Therefore, it is only necessary to consider feature extraction methods that can be used for RGB, RGB-D, and RGB-T data to simplify the model structure while ensuring performance.

\begin{figure}[t]
\centering
\includegraphics[width=\linewidth]{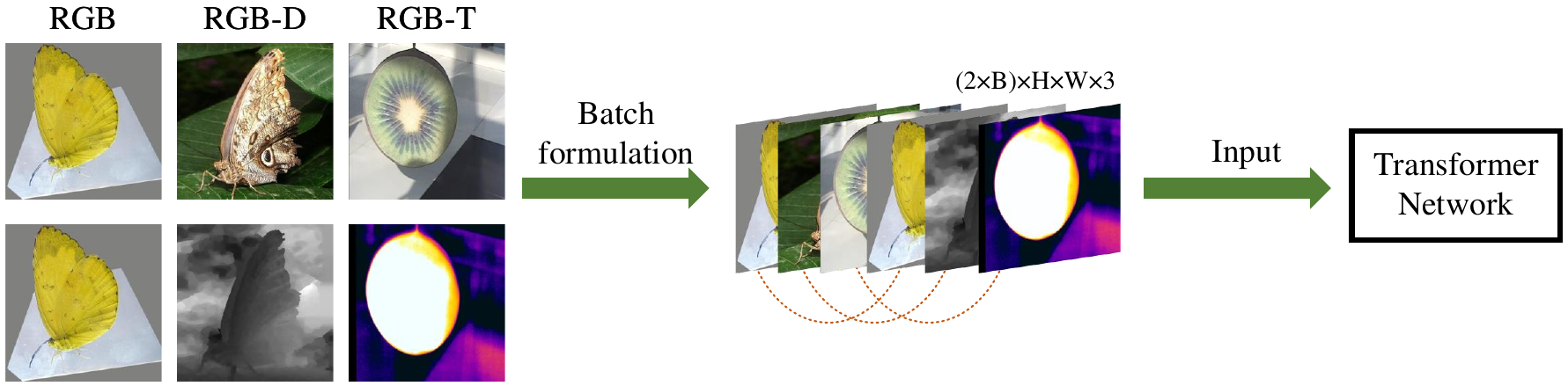}
\caption{Diagram of proposed model framework. The framework extracts RGB, RGB-D, and RGB-T data simultaneously with a single transformer network with shared weights.}
\label{trainstrategy}
\end{figure}

\begin{figure}[t]
\centering
\includegraphics[width=\linewidth]{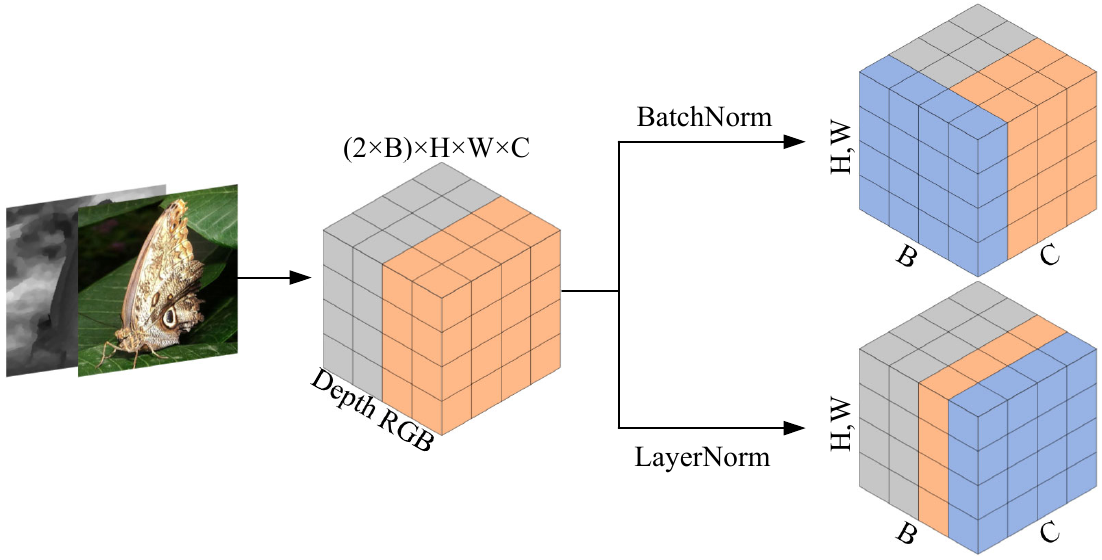}
\caption{BatchNorm and LayerNorm.}
\label{BNLN}
\end{figure}

In the training process, we consider the depth map or thermal map as a special kind of RGB map and merge them into an input batch with the RGB maps in an orderly manner, and extract the features with one transformer network as the backbone network, as shown in \cref{trainstrategy}. The network learns features from different modalities by sharing weights, and replaces the Batchnorm with the Layernorm thus avoiding the interference of batch normalisation in learning features from different modalities. As shown in \cref{BNLN}, a feature is obtained by concatenating RGB and Depth features in the Batch dimension. When BatchNorm is applied to this feature, the information from both modalities interferes with each other, whereas when LayerNorm is used, there is no interference. In summary, using a single transformer network with shared weights to extract multimodal features can prevent negative interference between the modalities, ensuring performance, and simplifying the model structure, effectively saving parameters. Therefore, this feature extraction method is suitable for single-modal RGB information as well as dual-modal RGB-D and RGB-T information.

We build a lightweight SOD model based on the proposed model framework, which is called AiOSOD due to its ability to perform saliency detection for all three data types in one model. Due to the large training sets of the three data,  AiOSOD is designed to be lightweight in order to validate the experimental results faster. AiOSOD employs a lightweight T2T-ViT-10 network \cite {yuan2021t2tvit} as the encoder and piggybacks on a lightweight decoder.

In conclusion, this paper has the following contributions:
\begin{itemize}
\item 
This work is the first time to consider all three (RGB, RGB-D, and RGB-T) saliency detection tasks all in one model. For the task of saliency detection of three different types of data, we introduces a novel model framework. This innovative model framework provides a unified solution for three types of data. It means that one weight file obtained through once training can be used universally for RGB, RGB-D, and RGB-T SOD. And the framework is successfully migrated in some 3D SOD models.
\item 
Proposed framework employs a single-stream transformer network with shared weight parameters for extracting multi-modal features. This feature extraction method ensures comprehensive training for all three data types while preventing interference between multi-modal features. This not only provides a unified solution for the detection of all three data types but also ensures precision in detecting these data types.
In comparison to models trained on only a single type of data, this framework achieves close performance and even make breakthroughs on some datasets. The effectiveness of this framework has been validated by our proposed model (AiOSOD) .
\item 
We propose a simple general model called AiOSOD to validate the proposed model framework. Thanks to the joint training of the three types of data and the feature extraction method, even though AiOSOD is a lightweight model, it achieves state-of-the-art performance in all three saliency detection tasks.
\end{itemize}

\section{Related work}
\label{sec:relatexwork}

CNN-based SOD methods have achieved many impressive results. In recent years, transformer networks have evolved in the field of computer vision, demonstrating excellent performance. The transformer structure is commonly used to model global remote dependencies between word sequences in machine translation tasks \cite{vaswani2017attention}. The self-attention mechanism is the core idea of transformer, which establishes associations between different positions in a sequence by calculating the correlation between a query and a key. Transformer captures long-distance dependencies by stacking multiple layers of self-attention layers. Vision Transformer (ViT) \cite{dosovitskiy2020vit} is the first application of the transformer structure to image classification tasks, being able to understand images from a holistic point of view. The ViT is suitable for a variety of vision tasks such as high-level classification and low-level dense prediction. Salient object detection belongs to the pixel-level dense prediction task. Therefore, more and more saliency detection models \cite{liu2021swinnet, liu2021vst, jia2022siatrans} have adopted the transformer structure to capture the global correlation information in images.

JL-DCF \cite{fu2021jldcf} first introduces siamese networks into RGB-D SOD, which employs a single convolutional network with shared parameters to extract common features of the two modalities. However, during training, the batch of JL-DCF can only be set to 1. If the batch size exceeds 1, its performance will decrease significantly. CSNet \cite{zhang2023csnet} employs a large kernel convolutional neural network as a siamese network to improve the performance. CCFENet \cite{liao2022ccfenet} introducs a cross-modal interaction module on the basis of weight sharing to enhance the siamese network's ability to learn features. However, neither of CSNet and CCFENet solve the problems of JL-DCF. In contrast, SiaTrans \cite{jia2022siatrans}  addresses this issue. It finds that multimodal features interfering with each other due to the effects of batch normalisation in multiple batches of training. SiaTrans adopts the transformer network as the siamese network, which conquers this drawback, makes full use of the computational power of the GPU and improves the training efficiency. Based on this property of the transformer, we can infer that the transformer network is suitable for the joint training of RGB, RGB-D, and RGB-T data.

\begin{figure*}[t]
\centering
\includegraphics[height=6.5cm]{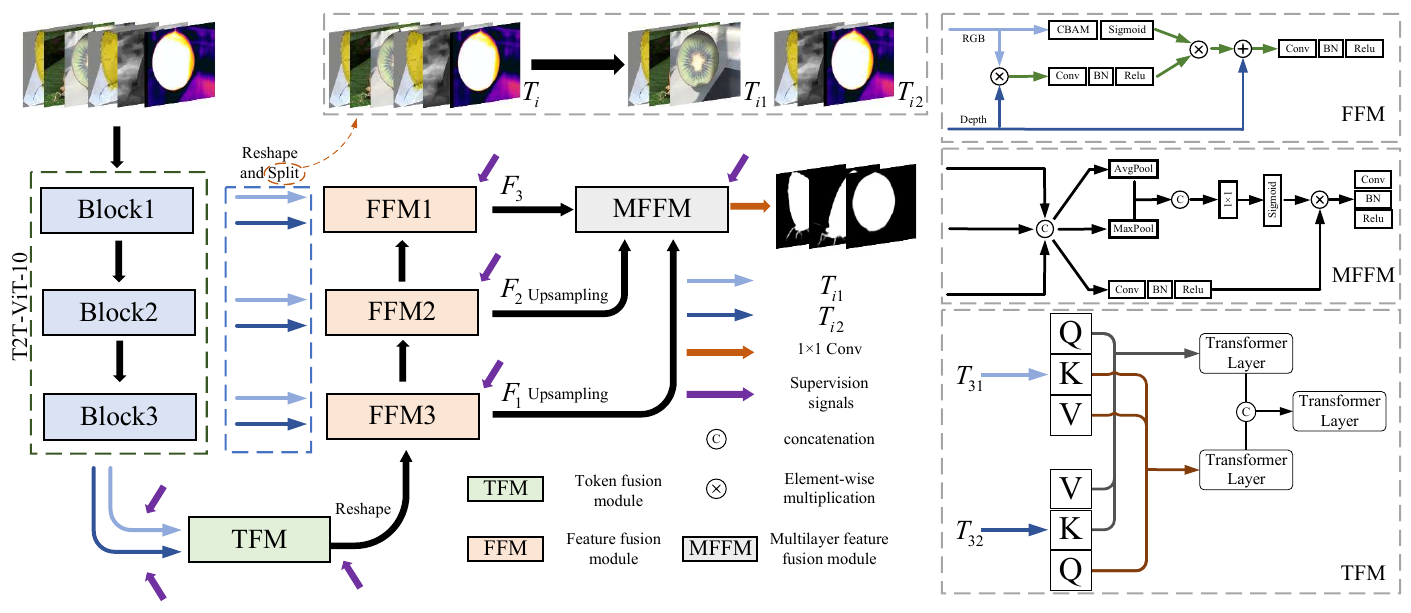}
\caption{Framework of our proposed AiOSOD.}
\label{fig3}
\end{figure*}

\section{Proposed model}

In order to achieve saliency detection across three types of data, we propose a lightweight AiOSOD model (as illustrated in \cref{fig3}). The model is mainly composed of three components: an encoder, a token fusion module (TFM), and a decoder. The encoder employs the T2T-ViT-10 network with 5.31M parameters. The token fusion module facilitates high-level cross-modal information integration, containing 0.29M parameters. The decoder, structured with convolutional architecture, consists of three feature fusion modules (FFM) and a multi-level feature fusion module (MFFM), totally containing  0.64M parameters.

\subsection{Encoder}
Considering the performance and parameters of networks like ViT \cite{dosovitskiy2020vit}, T2T-ViT \cite{yuan2021t2tvit}, PVTv2 \cite{wang2022pvt}, and Swin Transformer \cite{Swintransformer}, we adopt the T2T-ViT-10 in the work \cite{yuan2021t2tvit} as the backbone network for our AiOSOD model. T2T-ViT \cite{yuan2021t2tvit} is an improvement for the lack of local modeling capability of ViT \cite{dosovitskiy2020vit}. T2T-ViT incorporates the Tokens-to-Token (T2T) operation, whiche merges adjacent tokens into new tokens, effectively reducing token length and enabling local dependency modeling within images. 

The token sequence ${{T}_{0}}\in {{\mathbb{R}}^{l\times c}}$ derivs from input image ${I}\in {{\mathbb{R}}^{h\times w\times c}}$ after undergoing transformations, and serves as the input of the backbone network. Through successive transformer operations and Tokens-to-Token (T2T) processes, multilevel tokens sequences, namely ${{T}_{1}}$, ${{T}_{2}}$, and ${{T}_{3}}$, are generated.

\begin{equation}
{{T}_{i}}=Backbone\left( {{T}_{0}} \right),
\label{eq1}
\end{equation}
where $i = 1, 2, 3$. 
In order to reduce the parameters, ${{T}_{i}}$ is subsequently dimensionally transformed to be 64. After that ${{T}_{i}}$ needs to be splited in the order of concatenating of input  ${I}$ to obtain ${T}_{i1}$, ${T}_{i2}$ sequentially, as shown in \cref{fig3}.

\begin{equation}
{{T}_{i1}}, {{T}_{i2}}=Split\left( {{T}_{i}} \right).
\label{eq2}
\end{equation}
${{T}_{i1}}$ and ${{T}_{i2}}$ need to be reshaped into four-dimensional tensors to serve as inputs for the decoder consisting of a convolutional architecture. Additionally, ${{T}_{31}}$ and ${{T}_{32}}$ also serve as inputs of the tokens fusion module.

The tokens fusion module (TFM), as depicted in \cref{fig3}, is employed to integrate top-level tokens information, with a parameter count of 0.288M. Generally, with equivalent parameters, the computational complexity of a transformer block significantly exceeds that of a convolutional block. Therefore, considering computation and performance comprehensively, AiOSOD only fuses the top-level tokens of the two modalities. 

The “scaled dot-product attention” \cite{dosovitskiy2020vit} in the multi-head attention can be written as:
\begin{align}
Attention(Q,K,V) = \text{softmax}({{Q}K^T}/{\sqrt{{{d}_{k}}}}\;){V},
\label{eq3}
\end{align}
where $Q$ is Query, $K$ is Key, $V$ is Value, $d_k$ is the length of the Key vector. AiOSOD may have three types of inputs, which are (RGB, RGB) pairs, (RGB, depth) pairs, and (RGB, thermal) pairs. When the input consists of RGB and depth modalities, according to \cref{eq3}, TFM facilitates cross-modal interaction through the following process:
\begin{equation}
\small
\begin{split}
&\text{Attention}({{Q}_{R}},{{K}_{D}},{{V}_{D}})=\text{softmax}({{{Q}_{R}}K_{D}^{T}}/{\sqrt{{{d}_{k}}}}\;){{V}_{D}}, \\
&\text{Attention}({{Q}_{D}},{{K}_{R}},{{V}_{R}})=\text{softmax}({{{Q}_{D}}K_{R}^{T}}/{\sqrt{{{d}_{k}}}}\;){{V}_{R}}.
\end{split}
\label{eq4}
\end{equation}
Similarly, RGB-T flows can be processed according to \cref{eq4}. 
When the input involves two RGB flows, the Key and Value of these two RGB flows are the same.
TFM essentially reinforces self-attention for RGB tokens individually. 
Exactly because TFM can effectively handle different information  flow pairs, which is highly suitable for AiOSOD, we adopt TFM to fuse the top-level tokens.

\subsection{Decoder}
The decoder of the AiOSOD model consists of three feature fusion modules (FFM) and one multi-level feature fusion module (MFFM). The schematic diagrams of FFM and MFFM are provided in \cref{fig3}. The parameters of FFM is 0.071M, while that of MFFM is of 0.141M.

The FFM within the decoder serves not only as component modules that aggregate features from high to low levels in the decoder network but also effectively fuse features. Each FFM module is composed of a CBAM attention \cite{woo2018cbam} and two convolutional blocks. Through element-wise addition and multiplication, the FFM enhances salient features.

The multi-level feature fusion module (MFFM) is employed to fuse the three distinct-level features obtained from the FFMs, enhancing the accuracy of predictions. These features obtained from the FFMs, denoted as ${F}_1$, ${F}_2$, and ${F}_3$ respectively, are upsampled to the size of ($56\times56$), maintaining the same dimensions as ${F}_3$. Subsequently, the three features are fused using channel-wise attention and convolutional computations within the MFFM to achieve multi-level feature fusion.

\subsection{Implementation details}

\textit{\textbf{Training dataset.}} AiOSOD is trained jointly using three different types of data. The training dataset consists of the following subsets: the RGB dataset DUTS-TR \cite{DUTS} with 10,553 images, the RGB-T dataset VT5000 \cite{VT5000} with 2,500 image pairs, the RGB-D dataset NJUD \cite{njud} with 1,485 image pairs, NLPR \cite{NLPR} with 700 image pairs, and DUTLF-Depth \cite{DUTLF-D} with 800 image pairs.

\textit{\textbf{Loss function.}} We utilize the cross-entropy loss function:

\begin{equation}
\small
\mathcal{L}\left( P,G \right)=-\sum\limits_{i}{\left[ {{g}_{i}}log\left( {{p}_{i}} \right)+\left( 1-{{g}_{i}} \right)log\left( 1-{{p}_{i}} \right) \right]},
\label{eq5}
\end{equation}
where $P\in {{[0,1]}^{224\times 224}}$ and $G\in {{[0,1]}^{224\times 224}}$ represent the prediction map and ground truth (GT) map, respectively. ${{g}_{i}}\in G$, ${{p}_{i}}\in P$ represent individual pixel values.

\textit{\textbf{Training settings.}} Our proposed model is implemented by PyTorch \cite{paszke2019pytorch}, and trained on an RTX2080Ti (11GB). We resize each image to 256 × 256 pixels and then randomly crop 224 × 224 image regions as the model input.
We employ the Adam optimizer \cite{kingma2014adam} with an initial learning rate of 0.0001 and a batch size of 16. The training process includes 300,000 steps. The learning rate is reduced by a factor of 10 at steps 100,000 and 200,000.
\section{Benchmarking evaluation result}
In this study, we conduct benchmark tests on RGB datasets, RGB-D datasets, and RGB-T datasets. We compare AiOSOD against a total of 16 state-of-the-art (SOTA) models on these datasets.

\subsection{Evaluation metrics}
\textit{\textbf{MAE.}} 
The mean absolute error (MAE) \cite{mae-fm} represents the average absolute pixel difference between the saliency prediction map ($P$) and the ground truth map ($G$), and it is calculated using the following formula:
\begin{equation}
MAE=\frac{1}{W\times H}\sum\limits_{x=1}^{W}{\sum\limits_{y=1}^{H}{\lvert P\left( x,y \right)-G\left( x,y \right) \rvert}},
\label{eq6}
\end{equation}
where, $W$ and $H$ represent the width and height of the saliency map, respectively. A smaller error indicates a closer match between the prediction and the ground truth, thus indicating a more accurate prediction.

\textit{\textbf{F-Measure.}} 
The F-measure \cite{mae-fm} is a comprehensive performance metric calculated as the weighted harmonic mean of precision and recall. The formula is as follows:
\begin{equation}
F\text{-}measure = \frac{(1 + \beta^2) \times \text{Precision} \times \text{Recall}}{\beta^2 \times \text{Precision} + \text{Recall}},
\label{eq7}
\end{equation}
where $\beta$ is set to 0.3. We use the maximum F-measure as the evaluation metric, where a higher value indicates better prediction performance.

\textit{\textbf{S-Measure.}} 
The S-measure \cite{Smeasure} is particularly focused on evaluating the structural information within saliency maps and is considered closer to the human visual system than the F-measure. The formula for S-measure is expressed as:
\begin{equation}
S=\gamma {{S}_{0}}+\left( 1-\gamma  \right){{S}_{\gamma }},
\label{eq8}
\end{equation}
where ${{S}_{0}}$ and ${{S}_{\gamma }}$ denote region-aware and object-aware structural similarities, respectively. The parameter $\gamma$ is set to a default value of 0.5.
Absolutely, a higher S-measure value indicates more accurate predictions in terms of capturing structural information within the saliency maps.

\textit{\textbf{E-Measure.}} 
The E-measure \cite{Emeasure} is employed to quantify both global and local saliency differences and can be expressed as:
\begin{equation}
{{E}_{m}}=\frac{1}{W\times H}\sum\limits_{x=1}^{W}{\sum\limits_{y=1}^{H}{\phi \left( x,y \right)}},
\label{eq9}
\end{equation}
where \(\phi \left( \cdot  \right)\) represents the enhanced consistency matrix operation.
A larger value of $E_{m}$ indicates a more accurate prediction.

\textit{\textbf{FPS, Parameters, and FLOPs.}} 
In \cref{RGBRGBDRGBT,lightweighe}, we calculate the FPS, parameters, and FLOPs for these methods. The FPS calculation code is sourced from MobileSal \cite{MobileSal} and is tested on the same RTX2080Ti (11G) platform. The code for calculating the parameters and FLOPs is obtained from the Python library 'thop'.

\subsection{Comparison with SOTA RGB, RGB-D, and RGB-T  SOD models}

In \cref{RGBRGBDRGBT}, we compare our model (AiOSOD) with some SOTA models on RGB, RGB-D, and RGB-T datasets, respectively. RGB datasets include the testset of DUTS \cite{DUTS} (5019 images), DUT-OMORN \cite{DUT-OMORN} (5166 images), and ECSSD \cite{ECSSD} (1000 images). RGB SOD models include ICON (PAMI22 \cite{ICON}), ITSD (CVPR20 \cite{ITSD}),  RCSB (WACV22 \cite{RCSB}),  VST (ICCV21 \cite{liu2021vst}),  CII (TIP21 \cite{CII}), and CTDNet (ACM MM21 \cite{CTDNet}). RGB-D datasets include the testsets of DUTLF-Depth \cite{DUTLF-D} (400 pairs of images), NJUD \cite{njud} (500 pairs of images), NLPR \cite{NLPR} (300 pairs of images), and SIP \cite{SIP} (929 pairs of images). RGB-D SOD models include C2DFNet (IEEE TM \cite{C2DFNet}), CAVER (TIP23 \cite{CAVER}),  CCFENet (TCSVT22 \cite{liao2022ccfenet}),  JL-DCF (PAMI \cite{fu2021jldcf}),  SwinNet (TCSVT21 \cite{liu2021swinnet}), and VST. RGB-T datasets include the testset of VT5000 \cite{VT5000} (2500 pairs of images), along with VT821 \cite{VT821} (821 pairs of images), and VT1000 \cite{VT1000} (1000 pairs of images). RGB-T SOD models include  CCFENet, TNet (IEEE TM22 \cite{tnet}),  LSNet (TIP23 \cite{zhou2023lsnet}), CSRNet (TCSVT21 \cite{CSRNet}), SwinNet, and VST. And in \cref{lightweighe} our model is compared with three lightweight RGB-D models, LSNet, DFM-Net (ACM MM21 \cite{DFM-Net}), and MobileSal (PAMI21 \cite{MobileSal}).

\cref{tab:RGB,tab:RGBD,tab:RGBT}, present the comparative results on the RGB, RGB-D, and RGB-T datasets, respectively. Overall, benefiting from the proposed model framework, AiOSOD demonstrates a competitive advantage on RGB, RGB-D, and RGB-T datasets. It achieves excellent performance with low parameters, low computation requirements and high speed.
In terms of FPS, parameters and Flops, AiOSOD achieves the best results in RGB and RGB-D saliency detection methods, and it's second-best in RGB-T saliency detection. Regarding performance, AiOSOD delivers moderate performance on the RGB dataset. However, surprisingly, AiOSOD, as a lightweight model, achieves second-best performance across multiple datasets in RGB-D and RGB-T. The data in  \cref{RGBRGBDRGBT} demonstrates that AiOSOD is able to efficiently process RGB, RGB-D, and RGB-T data, maintaining a balance between performance, size and speed. 

\begin{table}[htbp]
\caption{Quantitative comparison of our model with other SOTA RGB, RGB-D, and RGB-T SOD methods on benchmark datasets. The best and second best results are highlighted in red and blue.}
\label{RGBRGBDRGBT}
\centering
    \begin{subtable}{\linewidth}
        \caption{RGB SOD methods}
        \resizebox{\linewidth}{!}{\begin{tabular}{cccccccc}
\hline
Method                     & ICON                          & ITSD     & RCSB                          & VST                           & CII                           & CTDNet                       & AiOSOD                       \\
Backbone                   & ResNet50                      & ResNet50 & ResNet50                      & T2T-ViT-t14                   & ResNet18                      & ResNet18                     & T2T-ViT-10                    \\
Speed(FPS)                 & 182                           & 90       & ——                            & 76                            & 293                           & {\color[HTML]{3531FF} 551}   & {\color[HTML]{FE0000} 780}    \\
Params(M)                  & 31.51                         & 24.87    & 31.91                         & 42.05                         & 11.34                         & {\color[HTML]{3531FF} 11.28} & {\color[HTML]{FE0000} 6.25}   \\
Flops(G)                   & 19.53                         & 14.87    & 265.12                        & 21.64                         & 14.89                         & {\color[HTML]{3531FF} 5.72}  & {\color[HTML]{FE0000} 2.04}   \\ \hline
\multicolumn{8}{c}{DUT-OMRON}                                                                                                                                                                                    \\ \hline
$Sm\uparrow$               & 0.8442                        & 0.8401   & 0.8350                        & {\color[HTML]{FE0000} 0.8501} & 0.8388                        & 0.8442   & {\color[HTML]{3531FF} 0.8447} \\
$F_\beta ^{max} \uparrow$  & {\color[HTML]{3531FF} 0.7985} & 0.7923   & 0.7727                        & {\color[HTML]{FE0000} 0.8001} & 0.7817                        & 0.7985   & 0.7890                        \\
$E_\phi ^{\max } \uparrow$ & {\color[HTML]{3531FF} 0.8842} & 0.8795   & 0.8659                        & {\color[HTML]{FE0000} 0.8878} & 0.8757                        & 0.8842   & 0.8823                        \\
$MAE\downarrow$            & 0.0569                        & 0.0608   & {\color[HTML]{FE0000} 0.0492} & 0.0579                        & {\color[HTML]{3531FF} 0.0538} & 0.0569   & 0.0549                        \\ \hline
\multicolumn{8}{c}{DUTS-TE}                                                                                                                                                                                      \\ \hline
$Sm\uparrow$               & {\color[HTML]{3531FF} 0.8886} & 0.8849   & 0.8808                        & {\color[HTML]{FE0000} 0.8961} & 0.8876                        & 0.8886   & 0.8821                        \\
$F_\beta ^{max} \uparrow$  & {\color[HTML]{3531FF} 0.8768} & 0.8680   & 0.8676                        & {\color[HTML]{FE0000} 0.8779} & 0.8696                        & 0.8768   & 0.8562                        \\
$E_\phi ^{\max } \uparrow$ & {\color[HTML]{3531FF} 0.9316} & 0.9294   & 0.9250                        & {\color[HTML]{FE0000} 0.9393} & 0.9290                        & 0.9316   & 0.9272                        \\
$MAE\downarrow$            & 0.0373                        & 0.0410   & {\color[HTML]{FE0000} 0.0350} & 0.0374                        & {\color[HTML]{3531FF} 0.0365} & 0.0373   & 0.0408                        \\ \hline
\multicolumn{8}{c}{ECSSD}                                                                                                                                                                                        \\ \hline
$Sm\uparrow$               & {\color[HTML]{3531FF} 0.9290} & 0.9248   & 0.9217                        & {\color[HTML]{FE0000} 0.9322} & 0.9261                        & 0.9290   & 0.9280                        \\
$F_\beta ^{max} \uparrow$  & {\color[HTML]{3531FF} 0.9433} & 0.9394   & 0.9355                        & {\color[HTML]{FE0000} 0.9442} & 0.9395                        & 0.9433   & 0.9387                        \\
$E_\phi ^{\max } \uparrow$ & 0.9603                        & 0.9589   & 0.9545                        & {\color[HTML]{FE0000} 0.9641} & 0.9562                        & 0.9603   & {\color[HTML]{3531FF} 0.9623} \\
$MAE\downarrow$            & {\color[HTML]{FE0000} 0.0318} & 0.0345   & 0.0335                        & {\color[HTML]{3531FF} 0.0329} & 0.0334                        & 0.0318   & 0.0339                        \\ \hline
\end{tabular}
} 
        \label{tab:RGB}
    \end{subtable}
    
    \begin{subtable}{\linewidth}
        \caption{RGB-D SOD methods}
        \resizebox{\linewidth}{!}{
        \begin{tabular}{cccccccc}
\hline
Method                     & C2DFNet                      & CAVER                         & CCFENet                       & JL-DCF                        & SwinNet                       & VST                           & AiOSOD                       \\
Backbone                   & ResNet50                     & ResNet50                      & ResNet50                      & ResNet50                      & Swin-B                        & T2T-ViT-t14                   & T2T-ViT-10                    \\
Speed(FPS)                 & {\color[HTML]{3531FF} 283}   & 144                           & 46                            & 17                            & 21                            & 58                            & {\color[HTML]{FE0000} 485}    \\
Params(M)                  & 45.31                        & 53.21                         & {\color[HTML]{3531FF} 27.33}  & 118.76                        & 189.57                        & 79.21                         & {\color[HTML]{FE0000} 6.25}   \\
Flops(G)                   & {\color[HTML]{3531FF} 10.32} & 20.36                         & 15.93                         & 787.88                        & 116.16                        & 28.95                         & {\color[HTML]{FE0000} 3.24}   \\ \hline
\multicolumn{8}{c}{DUTLF-Depth}                                                                                                                                                                                                                           \\ \hline
$Sm\uparrow$               & 0.9326                       & 0.9310                        & 0.9328                        & 0.8937                        & {\color[HTML]{FE0000} 0.9469} & 0.9426                        & {\color[HTML]{3531FF} 0.9455} \\
$F_\beta ^{max} \uparrow$  & 0.9440                       & 0.9404                        & 0.9445                        & 0.8920                        & {\color[HTML]{FE0000} 0.9579} & 0.9493                        & {\color[HTML]{3531FF} 0.9526} \\
$E_\phi ^{\max } \uparrow$ & 0.9638                       & 0.9639                        & 0.9636                        & 0.9283                        & {\color[HTML]{FE0000} 0.9752} & 0.9708                        & {\color[HTML]{3531FF} 0.9746} \\
$MAE\downarrow$            & 0.0252                       & 0.0283                        & 0.0263                        & 0.0488                        & {\color[HTML]{FE0000} 0.0226} & 0.0246                        & {\color[HTML]{3531FF} 0.0239} \\ \hline
\multicolumn{8}{c}{NJUD}                                                                                                                                                                                                                                  \\ \hline
$Sm\uparrow$               & 0.9079                       & 0.9203                        & 0.9165                        & 0.9104                        & {\color[HTML]{FE0000} 0.9255} & 0.9224                        & {\color[HTML]{3531FF} 0.9248} \\
$F_\beta ^{max} \uparrow$  & 0.9086                       & {\color[HTML]{3531FF} 0.9235} & 0.9210                        & 0.9119                        & {\color[HTML]{FE0000} 0.9283} & 0.9195                        & 0.9233                        \\
$E_\phi ^{\max } \uparrow$ & 0.9423                       & 0.9534                        & 0.9543                        & 0.9506                        & {\color[HTML]{FE0000} 0.9573} & 0.9510                        & {\color[HTML]{3531FF} 0.9567} \\
$MAE\downarrow$            & 0.0389                       & {\color[HTML]{FE0000} 0.0314}  & {\color[HTML]{3531FF} 0.0323}                       & 0.0410                        & {\color[HTML]{FE0000} 0.0314} & 0.0343                        & 0.0330                        \\ \hline
\multicolumn{8}{c}{NLPR}                                                                                                                                                                                                                                  \\ \hline
$Sm\uparrow$               & 0.9279                       & 0.9290                        & 0.9268                        & {\color[HTML]{3531FF} 0.9306}                       & 0.9296 & {\color[HTML]{FE0000} 0.9314} & 0.9273                        \\
$F_\beta ^{max} \uparrow$  & 0.9166                       & {\color[HTML]{FE0000} 0.9211} & 0.9184                        & 0.9182                        & 0.9170 & {\color[HTML]{3531FF} 0.9201}                       & 0.9115                        \\
$E_\phi ^{\max } \uparrow$ & 0.9605                       & {\color[HTML]{3531FF} 0.9638}                       & 0.9624                        & {\color[HTML]{FE0000} 0.9648} &   0.9624 & 0.9623                        & 0.9582                        \\
$MAE\downarrow$            & {\color[HTML]{3531FF} 0.0217}          & 0.0220                        & {\color[HTML]{FE0000} 0.0208} & 0.0221                        &  0.0225 & 0.0233                        & 0.0250                        \\ \hline
\multicolumn{8}{c}{SIP}                                                                                                                                                                                                                                   \\ \hline
$Sm\uparrow$               & 0.8715                       & 0.8934                        & 0.8816                        & 0.8852                        & 0.9009 & {\color[HTML]{3531FF} 0.9036}                        & {\color[HTML]{FE0000} 0.9069} \\
$F_\beta ^{max} \uparrow$  & 0.8770                       & 0.9064                        & 0.8977                        & 0.8935                        & 0.9122 & {\color[HTML]{3531FF} 0.9150}                        & {\color[HTML]{FE0000} 0.9216} \\
$E_\phi ^{\max } \uparrow$ & 0.9160                       & 0.9344                        & 0.9238                        & 0.9305                        &  0.9396 & {\color[HTML]{3531FF} 0.9439}                        & {\color[HTML]{FE0000} 0.9492} \\
$MAE\downarrow$            & 0.0529                       & 0.0424                        & 0.0473                        & 0.0490                        &  0.0409 & {\color[HTML]{3531FF} 0.0396}                        & {\color[HTML]{FE0000} 0.0375} \\ \hline
\end{tabular}}
        \label{tab:RGBD}
    \end{subtable}
    
    \begin{subtable}{\linewidth}
        \caption{RGB-T SOD methods}
        \resizebox{\linewidth}{!}{
        \begin{tabular}{cccccccc}
\hline
Method                     & CCFENet                       & TNet                          & SwinNet                       & VST         & LSNet                       & CSRNet   & AiOSOD                       \\
Backbone                   & ResNet50                      & ResNet50                      & Swin-B                        & T2T-ViT-t14 & MobileNetV2                 & ESPNetv2 & T2T-ViT-10                    \\
Speed(FPS)                 & 46                            & 67                            & 21                            & 58          & {\color[HTML]{FE0000} 896}  & ——       & {\color[HTML]{3531FF} 485}    \\
Params(M)                  & 27.33                         & 83.35                         & 189.57                        & 79.21       & {\color[HTML]{FE0000} 4.35} & ——       & {\color[HTML]{3531FF} 6.25}   \\
Flops(G)                   & 15.93                         & 51.13                         & 116.16                        & 28.95       & {\color[HTML]{FE0000} 1.15} & 4.20     & {\color[HTML]{3531FF} 3.24}   \\ \hline
\multicolumn{8}{c}{VT800}                                                                                                                                                                              \\ \hline
$Sm\uparrow$               & {\color[HTML]{3531FF} 0.8996} & 0.8989                        & 0.8935                        & 0.8832      & 0.8786                      & 0.8847   & {\color[HTML]{FE0000} 0.9049} \\
$F_\beta ^{max} \uparrow$  & 0.8819                        & {\color[HTML]{FE0000} 0.8884} & 0.8707                        & 0.8541      & 0.8448                      & 0.8579   & {\color[HTML]{3531FF} 0.8821} \\
$E_\phi ^{\max } \uparrow$ & 0.9367                        & {\color[HTML]{FE0000} 0.9382} & 0.9288                        & 0.9172      & 0.9205                      & 0.9226   & {\color[HTML]{3531FF} 0.9372} \\
$MAE\downarrow$            & {\color[HTML]{FE0000} 0.0273} & 0.0302                        & 0.0334                        & 0.0412      & 0.0332                      & 0.0376   & {\color[HTML]{3531FF} 0.0283} \\ \hline
\multicolumn{8}{c}{VT1000}                                                                                                                                                                             \\ \hline
$Sm\uparrow$               & 0.9341                        & 0.9286                        & {\color[HTML]{3531FF} 0.9360} & 0.9329      & 0.9256                      & 0.9183   & {\color[HTML]{FE0000} 0.9410} \\
$F_\beta ^{max} \uparrow$  & 0.9336                        & 0.9296                        & {\color[HTML]{3531FF} 0.9392} & 0.9314      & 0.9216                      & 0.9083   & {\color[HTML]{FE0000} 0.9405} \\
$E_\phi ^{\max } \uparrow$ & 0.9684                        & 0.9662                        & {\color[HTML]{3531FF} 0.9727} & 0.9705      & 0.9626                      & 0.9525   & {\color[HTML]{FE0000} 0.9741} \\
$MAE\downarrow$            & {\color[HTML]{3531FF} 0.0182} & 0.0212                        & {\color[HTML]{FE0000} 0.0179} & 0.0211      & 0.0227                      & 0.0242   & 0.0202                        \\ \hline
\multicolumn{8}{c}{VT5000}                                                                                                                                                                             \\ \hline
$Sm\uparrow$               & {\color[HTML]{3531FF} 0.8960} & 0.8952                        & {\color[HTML]{FE0000} 0.9046} & 0.8870      & 0.8774                      & 0.8677   & 0.8958                        \\
$F_\beta ^{max} \uparrow$  & 0.8802                        & {\color[HTML]{3531FF} 0.8809} & {\color[HTML]{FE0000} 0.8920} & 0.8610      & 0.8499                      & 0.8372   & 0.8750                        \\
$E_\phi ^{\max } \uparrow$ & {\color[HTML]{3531FF} 0.9389} & 0.9374                        & {\color[HTML]{FE0000} 0.9481} & 0.9286      & 0.9240                      & 0.9138   & 0.9375                        \\
$MAE\downarrow$            & {\color[HTML]{3531FF} 0.0304} & 0.0328                        & {\color[HTML]{FE0000} 0.0290} & 0.0383      & 0.0370                      & 0.0416   & 0.0346                        \\ \hline
\end{tabular}}
        \label{tab:RGBT}
    \end{subtable}
\end{table}

\begin{table}[t]
\caption{Comparison with lightweight RGB-D SOD models.}
\label{lightweighe}
\centering
\resizebox{\linewidth}{!}{
\begin{tabular}{cccccccc}
\hline
Method                     & DFM-Net                       & LSNet                         & MobileSal                     & AiOSOD                       \\
Backbone                   & MobileNetV2                   & MobileNetV2                   & MobileNetV2                   & T2T-ViT-10                    \\
Speed(FPS)                 & {\color[HTML]{3531FF} 692}    & {\color[HTML]{FE0000} 896}    & 459                           & 485                           \\
Params(M)                  & {\color[HTML]{FE0000} 3.84}   & {\color[HTML]{3531FF} 4.35}   & 6.24                          & 6.25                          \\
Flops(G)                   & 2.68                          & {\color[HTML]{FE0000} 1.15}   & {\color[HTML]{3531FF} 1.51}   & 3.24                          \\ \hline
\multicolumn{5}{c}{NJUD}                                                                                                                                   \\ \hline
$Sm\uparrow$               & 0.9072                        & {\color[HTML]{3531FF} 0.9111} & 0.9097                        & {\color[HTML]{FE0000} 0.9248} \\
$F_\beta ^{max} \uparrow$  & 0.9130                        & {\color[HTML]{3531FF} 0.9144} & 0.9117                        & {\color[HTML]{FE0000} 0.9233} \\
$E_\phi ^{\max } \uparrow$ & {\color[HTML]{3531FF} 0.9525} & 0.9498                        & 0.9502                        & {\color[HTML]{FE0000} 0.9567} \\
$MAE\downarrow$            & 0.0428                        & 0.0386                        & {\color[HTML]{3531FF} 0.0370} & {\color[HTML]{FE0000} 0.0330} \\ \hline
\multicolumn{5}{c}{SIP}                                                                                                                                    \\ \hline
$Sm\uparrow$               & 0.8831                        & {\color[HTML]{3531FF} 0.8861} & 0.8732                        & {\color[HTML]{FE0000} 0.9069} \\
$F_\beta ^{max} \uparrow$  & 0.8873                        & {\color[HTML]{3531FF} 0.8952} & 0.8795                        & {\color[HTML]{FE0000} 0.9216} \\
$E_\phi ^{\max } \uparrow$ & 0.9259                        & {\color[HTML]{3531FF} 0.9306} & 0.9162                        & {\color[HTML]{FE0000} 0.9492} \\
$MAE\downarrow$            & 0.0507                        & {\color[HTML]{3531FF} 0.0496} & 0.0528                        & {\color[HTML]{FE0000} 0.0375} \\ \hline
\end{tabular}
}
\end{table}

\cref{lightweighe} shows the results of the comparison between AiOSOD and the three lightweight RGB-D SOD models on NJUD and SIP datasets. Compared to the lightweight models DFM-Net, LSNet, and MobileSal,  AiOSOD's parameters and FLOPs are slightly higher but it maintains the strongest performance. AiOSOD can process RGB-D images at 485 FPS, slower than DFM-Net and LSNet, and faster than MobileSal. It can be seen that AiOSOD is able to maintain high speed while maintaining a high accuracy.

\begin{figure}[htbp]
    \begin{subfigure}{\linewidth}
	\centering
	\includegraphics[width=\linewidth]{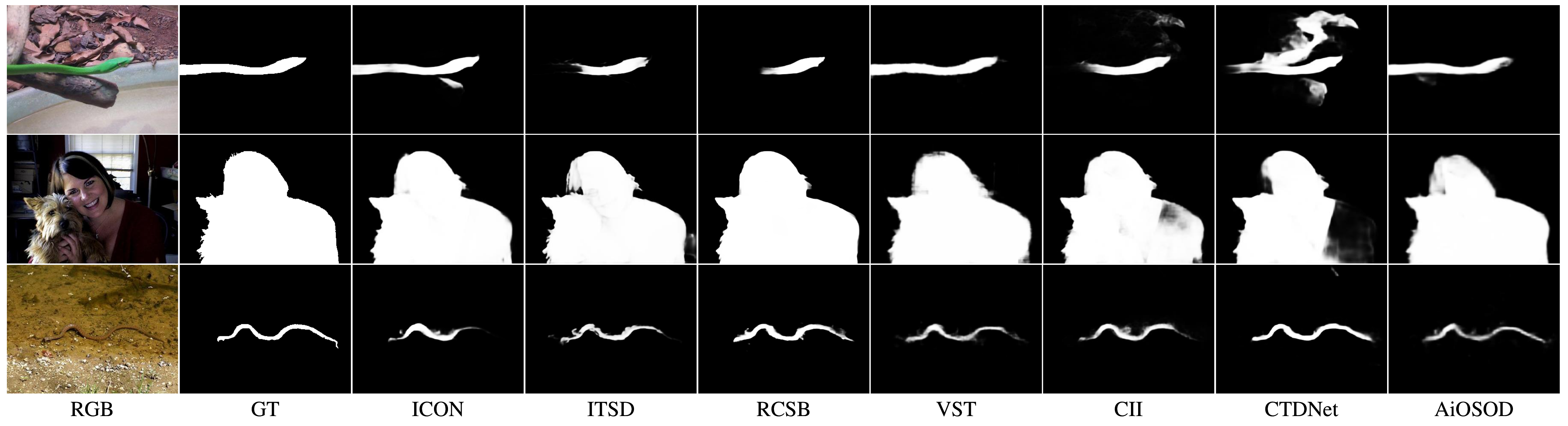}
	\caption{RGB SOD methods}
        \label{RGBIMAGE}
    \end{subfigure}
    
    \begin{subfigure}{\linewidth}
	\centering
	\includegraphics[width=\linewidth]{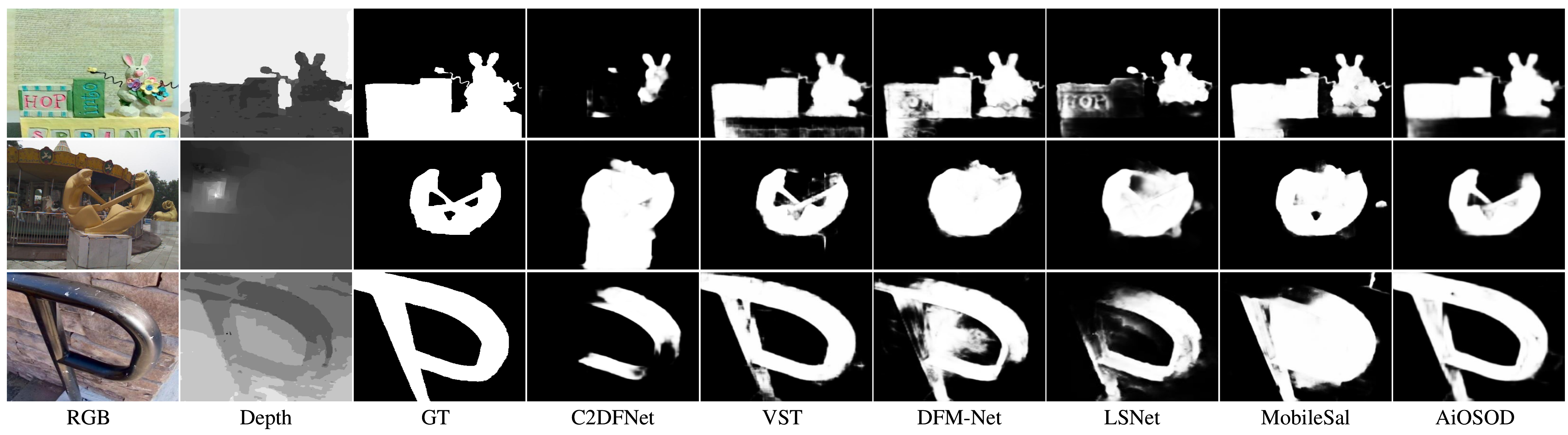}
	\caption{RGB-D SOD methods}
        \label{RGBDIMAGE}
    \end{subfigure}    
    
    \begin{subfigure}{\linewidth}
	\centering
	\includegraphics[width=\linewidth]{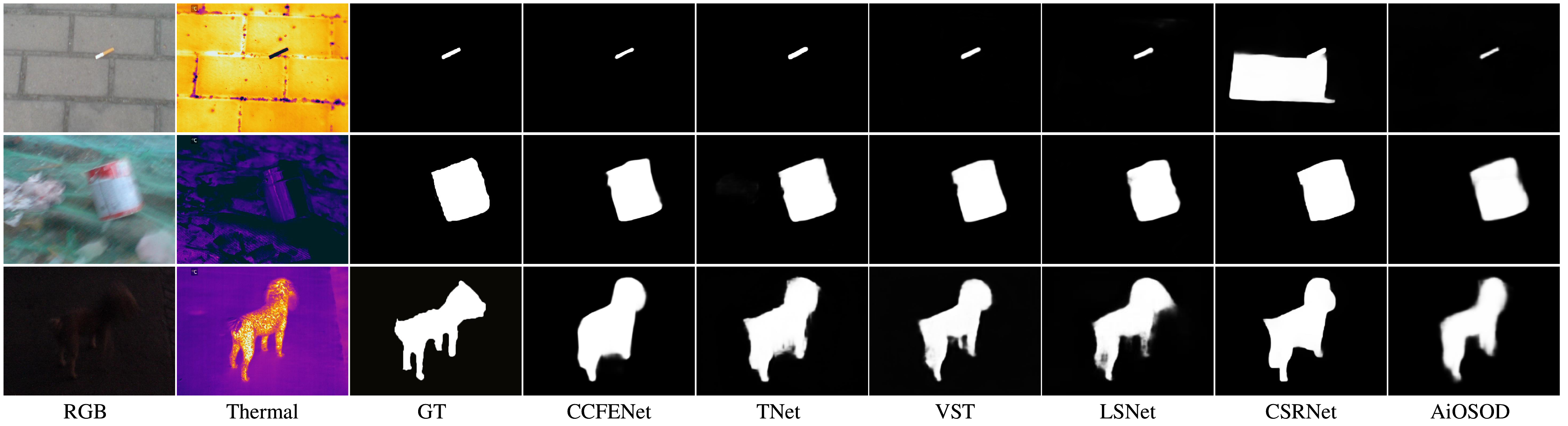}
	\caption{RGB-T SOD methods}
        \label{RGBTIMAGE}
    \end{subfigure}
\caption{Qualitative comparison with SOTA RGB, RGB-D, and RGB-T SOD methods.}
\label{Qualitative}
\end{figure}

As shown in \cref{Qualitative}, we demonstrate the generation of saliency maps in various challenging scenarios. \cref{RGBIMAGE} shows representative RGB images where AiOSOD consistently generates high-quality predictions, even in challenging conditions such as blurry boundaries, low contrast and small objects. Some representative RGB-D examples are shown in \cref{RGBDIMAGE}. AiOSOD effectively utilizes depth information to enhance saliency detection, maintaining good performance even when the depth maps are blurry. \cref{RGBTIMAGE} shows several representative RGB-T images.

\section{Discussion}

\subsection{Function of proposed components}

\cref{tab6} presents the results of ablation experiments on there datasets concerning the proposed components. In these experiments, we start with a baseline model that adopts a classical U-shaped architecture, utilizing the TiT-ViT-10 network as the encoder and three dual convolution layers as the decoder. The baseline model also follows our proposed model framework and integrates cross-modal information through element-wise addition. From the experimental outcomes, it is evident that the baseline model performs well on the RGB, RGB-D, and RGB-T datasets, thanks to our proposed framework. 

\begin{table}[t]
\renewcommand{\arraystretch}{1}
\caption{Quantitative evaluation results of ablation experiments for proposed components.}
\Large
\resizebox{\linewidth}{!}{
\begin{tabular}{l|c|cc|cc|cc}
\hline
                 & \multirow{2}{*}{Params(M)} & \multicolumn{2}{c|}{DUTLF-Depth}            & \multicolumn{2}{c|}{VT821}                  & \multicolumn{2}{c}{ECSSD}                   \\
                 &                            & $MAE\downarrow$ & $F_\beta ^{max} \uparrow$ & $MAE\downarrow$ & $F_\beta ^{max} \uparrow$ & $MAE\downarrow$ & $F_\beta ^{max} \uparrow$ \\ \hline
Baseline         & 5.80                       & 0.0308          & 0.9407                    & 0.0366          & 0.8685                    & 0.0356          & 0.9369                    \\
Base+TFM         & 6.10                       & 0.0282          & 0.9434                    & 0.0335          & 0.8788                    & 0.0350          & 0.9378                    \\
Base+TFM+FFM     & 6.10                       & 0.0264          & 0.9501                    & 0.0321          & 0.8697                    & 0.0370          & 0.9321                    \\
Base+TFM+FFM+MLF & 6.25                       & \textbf{0.0239} & \textbf{0.9526}           & \textbf{0.0283} & \textbf{0.8821}           & \textbf{0.0339} & \textbf{0.9387}           \\ \hline
\end{tabular}
}
\label{tab6}
\end{table}

Subsequently, we add a tokens fusion module on the top level of the baseline model (Base+TFM model), which increases 0.3M parameters. Compared to the baseline model, Base+TFM model demonstrates performance improvements across all there datasets, affirming the effectiveness of the tokens fusion module.

Building upon Base+TFM model, we further introduce there feature fusion modules as model decoder 
 (Base+TFM+FFM model). 
With very few additional parameters compared to the Base+TFM model, the Base+TFM+FFM model improves the detection of RGB-D and RGB-T data.
However, due to the FFM's primary focus on cross-modal feature fusion, its performance on the RGB dataset lags behind Base+TFM and baseline models.

In the final model (Base+TFM+FFM+MFFM model), we introduce a multi-level feature fusion module, namely AiOSOD. With this module, the performance of the model is successfully improved, thus improving all the metrics for the three types of datasets. Overall, these ablation experiment results confirm the effectiveness of our proposed model framework and individual components, showcasing their performance-enhancing effects across diverse datasets.

\subsection{Function of proposed model framework}
This subsection conducts experiments with JL-DCF \cite{fu2021jldcf} and VST \cite{liu2021vst} to explore the impact of extracting features with weight sharing in both convolutional and transformer networks. 
JL-DCF is the first RGB-D SOD model that extracts RGB and depth features through a convolutional network with shared weights, but its batch can only be set to 1. As shown in \cref{BNLN}, the convolutional network uses BatchNorm to normalize the RGB and depth features in the batch dimension. The experimental results of JL-DCF are shown in \cref{jldcfvst}.
When batch size is 1, only one pair of RGB and depth features is normalized, which has little effect on the prediction results. However, when the batch size exceeds 1, more than one pair of RGB and depth features are normalized, which affect each other and cause performance degradation.
For the VST model, using a single transformer network with weight sharing to extract multi-modal features produces results comparable to the original model and is not limited by batch size. 
These experiments confirm that convolutional networks are not suited to providing a unified solution for three data types and achieving optimal performance. Therefore, using a transformer network as the feature extraction network in our proposed  model framework is undoubtedly the best choice to meet the task requirements.

\begin{table}[t]
\centering
\caption{Impact of backbone network weight sharing on the prediction results of two RGB-D SOD models, JL-DCF and VST.}
\resizebox{\linewidth}{!}{

\begin{tabular}{c|cc|cc|cc|cc}
\hline
                        & Size & \multirow{2}{*}{Batch} & \multicolumn{2}{c|}{NJUD}                   & \multicolumn{2}{c|}{SIP}                    & \multicolumn{2}{c}{NLPR}                    \\
                        & (Mb) &                        & $MAE\downarrow$ & $F_\beta ^{max} \uparrow$ & $MAE\downarrow$ & $F_\beta ^{max} \uparrow$ & $MAE\downarrow$ & $F_\beta ^{max} \uparrow$ \\ \hline
\multirow{2}{*}{JL-DCF} & 475  & 1                      & \textbf{0.0454} & 0.8975                    & \textbf{0.0491} & \textbf{0.8890}           & \textbf{0.0216} & \textbf{0.9176}           \\
                        & 475  & 6                      & 0.0511          & \textbf{0.9038}           & 0.0614          & 0.8604                    & 0.0292          & 0.9030                    \\ \hline
\multirow{2}{*}{VST}    & 320  & 6                      & 0.0351          & 0.9195                    & 0.0403          & 0.9150                    & 0.0236          & 0.9201                    \\
                        & 238  & 6                      & \textbf{0.0324} & \textbf{0.9275}           & \textbf{0.0395} & \textbf{0.9149}           & \textbf{0.0234} & \textbf{0.9232}           \\ \hline
\end{tabular}
}
\label{jldcfvst}
\end{table}

\cref{backbone} presents the results of AiOSOD models with different backbone networks (T2T-ViT-10 \cite{yuan2021t2tvit}, PVTv2-B0 \cite{wang2022pvt}, and Swin Tiny \cite{Swintransformer}), including trained jointly on the RGB, RGB-D, and RGB-T datasets, as well as models trained separately on each of these datasets. 
Among these models based on PVTv2-B0 and Swin Tiny architectures, the original model structure is retained, and an additional FFM module is introduced. Of course, the number of convolutional kernels in each layer has also been adjusted based on the backbone network. Experimental results on T2T-ViT-10, PVTv2-B0, and Swin Tiny consistently show that joint training using all three data types produces better results compared to training on a single data type. This outcome can be attributed to the effectiveness of the proposed model framework, which not only reduces interference between different data types but also significantly increases the effective training sample size, thereby enhancing predictive performance. The data in \cref{backbone} confirms that the proposed model framework not only offers a unified solution for all three data types but also improves performance metrics across various testing datasets.

\begin{table}[ht]
\caption{Comparison of training results for AiOSOD models using different backbone networks. JT means joint training on RGB, RGB-D, and RGB-T datasets.}
\label{backbone}
\centering
    \begin{subtable}{\linewidth}
        \caption{RGB Datasets}
        \resizebox{\linewidth}{!}{
        \begin{tabular}{ccccccc}
\hline
\multicolumn{1}{c|}{Backbone}                  & \multicolumn{2}{c|}{T2T-ViT-10}               & \multicolumn{2}{c|}{PVTv2-B0}                 & \multicolumn{2}{c}{Swin Tiny} \\
\multicolumn{1}{c|}{Train dataset}             & RGB    & \multicolumn{1}{c|}{JT}              & RGB    & \multicolumn{1}{c|}{JT}              & RGB       & JT                \\ \hline
\multicolumn{7}{c}{DUTS-TE}                                                                                                                                                    \\ \hline
\multicolumn{1}{c|}{$Sm\uparrow$}              & 0.8786 & \multicolumn{1}{c|}{\textbf{0.8821}} & 0.8790 & \multicolumn{1}{c|}{\textbf{0.8797}} & 0.8905    & \textbf{0.8931}   \\
\multicolumn{1}{c|}{$F_\beta^{max} \uparrow$}  & 0.8521 & \multicolumn{1}{c|}{\textbf{0.8562}} & 0.8519 & \multicolumn{1}{c|}{\textbf{0.8539}} & 0.8659    & \textbf{0.8701}   \\
\multicolumn{1}{c|}{$E_\phi^{\max } \uparrow$} & 0.9234 & \multicolumn{1}{c|}{\textbf{0.9272}} & 0.9239 & \multicolumn{1}{c|}{\textbf{0.9262}} & 0.9329    & \textbf{0.9365}   \\
\multicolumn{1}{c|}{$MAE\downarrow$}           & 0.0418 & \multicolumn{1}{c|}{\textbf{0.0408}} & 0.0418 & \multicolumn{1}{c|}{\textbf{0.0412}} & 0.0387    & \textbf{0.0381}   \\ \hline
\multicolumn{7}{c}{DUT-OMRON}                                                                                                                                                  \\ \hline
\multicolumn{1}{c|}{$Sm\uparrow$}              & 0.8365 & \multicolumn{1}{c|}{\textbf{0.8447}} & 0.8383 & \multicolumn{1}{c|}{\textbf{0.8472}} & 0.8445    & \textbf{0.8535}   \\
\multicolumn{1}{c|}{$F_\beta^{max} \uparrow$}  & 0.7773 & \multicolumn{1}{c|}{\textbf{0.7890}} & 0.7772 & \multicolumn{1}{c|}{\textbf{0.7922}} & 0.7854    & \textbf{0.8018}   \\
\multicolumn{1}{c|}{$E_\phi^{\max } \uparrow$} & 0.8728 & \multicolumn{1}{c|}{\textbf{0.8823}} & 0.8746 & \multicolumn{1}{c|}{\textbf{0.8875}} & 0.8788    & \textbf{0.8928}   \\
\multicolumn{1}{c|}{$MAE\downarrow$}           & 0.0599 & \multicolumn{1}{c|}{\textbf{0.0549}} & 0.0566 & \multicolumn{1}{c|}{\textbf{0.0531}} & 0.0548    & \textbf{0.0528}   \\ \hline
\end{tabular}
        \label{tab:backbone-RGB}} 
    \end{subtable}
    
    \begin{subtable}{\linewidth}
        \caption{RGB-D Datasets}
        \resizebox{\linewidth}{!}{
        \begin{tabular}{ccccccc}
\hline
\multicolumn{1}{c|}{Backbone}                  & \multicolumn{2}{c|}{T2T-ViT-10}               & \multicolumn{2}{c|}{PVTv2-B0}                 & \multicolumn{2}{c}{Swin Tiny} \\
\multicolumn{1}{c|}{Train dataset}             & RGB-D  & \multicolumn{1}{c|}{JT}              & RGB-D  & \multicolumn{1}{c|}{JT}              & RGB-D     & JT                \\ \hline
\multicolumn{7}{c}{NJUD}                                                                                                                                                       \\ \hline
\multicolumn{1}{c|}{$Sm\uparrow$}              & 0.9136 & \multicolumn{1}{c|}{\textbf{0.9248}} & 0.9133 & \multicolumn{1}{c|}{\textbf{0.9247}} & 0.9207    & \textbf{0.9309}   \\
\multicolumn{1}{c|}{$F_\beta^{max} \uparrow$}  & 0.9093 & \multicolumn{1}{c|}{\textbf{0.9233}} & 0.9092 & \multicolumn{1}{c|}{\textbf{0.9241}} & 0.9209    & \textbf{0.9316}   \\
\multicolumn{1}{c|}{$E_\phi^{\max } \uparrow$} & 0.9474 & \multicolumn{1}{c|}{\textbf{0.9567}} & 0.9459 & \multicolumn{1}{c|}{\textbf{0.9552}} & 0.9549    & \textbf{0.9619}   \\
\multicolumn{1}{c|}{$MAE\downarrow$}           & 0.0386 & \multicolumn{1}{c|}{\textbf{0.0330}} & 0.0395 & \multicolumn{1}{c|}{\textbf{0.0343}} & 0.0366    & \textbf{0.0308}   \\ \hline
\multicolumn{7}{c}{SIP}                                                                                                                                                        \\ \hline
\multicolumn{1}{c|}{$Sm\uparrow$}              & 0.8971 & \multicolumn{1}{c|}{\textbf{0.9069}} & 0.8876 & \multicolumn{1}{c|}{\textbf{0.9001}} & 0.8949    & \textbf{0.9112}   \\
\multicolumn{1}{c|}{$F_\beta^{max} \uparrow$}  & 0.9082 & \multicolumn{1}{c|}{\textbf{0.9216}} & 0.8965 & \multicolumn{1}{c|}{\textbf{0.9116}} & 0.9043    & \textbf{0.9243}   \\
\multicolumn{1}{c|}{$E_\phi^{\max } \uparrow$} & 0.9403 & \multicolumn{1}{c|}{\textbf{0.9492}} & 0.9349 & \multicolumn{1}{c|}{\textbf{0.9427}} & 0.9403    & \textbf{0.9515}   \\
\multicolumn{1}{c|}{$MAE\downarrow$}           & 0.0423 & \multicolumn{1}{c|}{\textbf{0.0375}} & 0.0471 & \multicolumn{1}{c|}{\textbf{0.0414}} & 0.0439    & \textbf{0.0370}   \\ \hline
\end{tabular}
        \label{tab:backbone-RGBD}} 
    \end{subtable}

    \begin{subtable}{\linewidth}
        \caption{RGB-T Datasets}
        \resizebox{\linewidth}{!}{
        \begin{tabular}{ccccccc}
\hline
\multicolumn{1}{c|}{Backbone}                  & \multicolumn{2}{c|}{T2T-ViT-10}               & \multicolumn{2}{c|}{PVTv2-B0}                 & \multicolumn{2}{c}{Swin Tiny} \\
\multicolumn{1}{c|}{Train dataset}             & RGB-T  & \multicolumn{1}{c|}{JT}              & RGB-T  & \multicolumn{1}{c|}{JT}              & RGB-T     & JT                \\ \hline
\multicolumn{7}{c}{VT1000}                                                                                                                                                     \\ \hline
\multicolumn{1}{c|}{$Sm\uparrow$}              & 0.9263 & \multicolumn{1}{c|}{\textbf{0.9410}} & 0.9297 & \multicolumn{1}{c|}{\textbf{0.9388}} & 0.9309    & \textbf{0.9428}   \\
\multicolumn{1}{c|}{$F_\beta^{max} \uparrow$}  & 0.9251 & \multicolumn{1}{c|}{\textbf{0.9405}} & 0.9312 & \multicolumn{1}{c|}{\textbf{0.9406}} & 0.9293    & \textbf{0.9447}   \\
\multicolumn{1}{c|}{$E_\phi^{\max } \uparrow$} & 0.9654 & \multicolumn{1}{c|}{\textbf{0.9741}} & 0.9694 & \multicolumn{1}{c|}{\textbf{0.9745}} & 0.9699    & \textbf{0.9775}   \\
\multicolumn{1}{c|}{$MAE\downarrow$}           & 0.0240 & \multicolumn{1}{c|}{\textbf{0.0202}} & 0.0240 & \multicolumn{1}{c|}{\textbf{0.0202}} & 0.0225    & \textbf{0.0187}   \\ \hline
\multicolumn{7}{c}{VT5000}                                                                                                                                                     \\ \hline
\multicolumn{1}{c|}{$Sm\uparrow$}              & 0.8803 & \multicolumn{1}{c|}{\textbf{0.8958}} & 0.8793 & \multicolumn{1}{c|}{\textbf{0.8940}} & 0.8851    & \textbf{0.9034}   \\
\multicolumn{1}{c|}{$F_\beta^{max} \uparrow$}  & 0.8467 & \multicolumn{1}{c|}{\textbf{0.8750}} & 0.8508 & \multicolumn{1}{c|}{\textbf{0.8753}} & 0.8586    & \textbf{0.8880}   \\
\multicolumn{1}{c|}{$E_\phi^{\max } \uparrow$} & 0.9238 & \multicolumn{1}{c|}{\textbf{0.9375}} & 0.9251 & \multicolumn{1}{c|}{\textbf{0.9379}} & 0.9338    & \textbf{0.9457}   \\
\multicolumn{1}{c|}{$MAE\downarrow$}           & 0.0404 & \multicolumn{1}{c|}{\textbf{0.0346}} & 0.0414 & \multicolumn{1}{c|}{\textbf{0.0364}} & 0.0377    & \textbf{0.0325}   \\ \hline
\end{tabular}
        \label{tab:backbone-RGBT}} 
    \end{subtable}

\end{table}

\subsection{Applying proposed model framework to other models}

\begin{table}[t]
\caption{Applying proposed model framework to other models results. ST means training on a single type of data. JT means joint training on RGB, RGB-D, and RGB-T datasets. Dual-s and Single-s indicate that the model employs a dual-stream backbone network  or a single-stream backbone network.}
\label{tab:VSTSwinNetLSNet}
\centering
    \begin{subtable}{\linewidth}
        \caption{RGB  Datasets}
        \resizebox{\linewidth}{!}{
\begin{tabular}{cc|cccc|cccc}
\hline
\multicolumn{2}{c|}{\multirow{2}{*}{Model}} & \multicolumn{4}{c|}{DUTS-TE}                                                                  & \multicolumn{4}{c}{DUT-OMRON}                                                                 \\
\multicolumn{2}{c|}{}                       & $Sm\uparrow$    & $F_\beta ^{max} \uparrow$ & $E_\phi ^{\max } \uparrow$ & $MAE\downarrow$ & $Sm\uparrow$    & $F_\beta ^{max} \uparrow$ & $E_\phi ^{\max } \uparrow$ & $MAE\downarrow$ \\ \hline
\multirow{3}{*}{VST}        & Single-s-ST     & 0.8961          & 0.8779                    & \textbf{0.9393}            & 0.0374          & 0.8501          & 0.8001                    & 0.8878                     & 0.0579          \\
                            & Dual-s-JT     & \textbf{0.8993} & \textbf{0.8819}           & 0.9390                     & \textbf{0.0357} & \textbf{0.8597} & 0.8152                    & 0.8962                     & \textbf{0.0535} \\
                            & Single-s-JT   & 0.8974          & 0.8806                    & 0.9390                     & 0.0364          & \textbf{0.8597} & \textbf{0.8183}           & \textbf{0.8973}            & 0.0544          \\ \hline
\multirow{3}{*}{SwinNet}    & Dual-s-ST     & ——              & ——                        & ——                         & ——              & ——              & ——                        & ——                         & ——              \\
                            & Dual-s-JT     & 0.9025          & 0.8873                    & 0.9403                     & 0.0316          & 0.8675          & 0.8245                    & 0.9048                     & \textbf{0.0438} \\
                            & Single-s-JT   & \textbf{0.9051} & \textbf{0.8915}           & \textbf{0.9437}            & \textbf{0.0312} & \textbf{0.8683} & \textbf{0.8268}           & \textbf{0.9068}            & 0.0458          \\ \hline
\multirow{3}{*}{LSNet}      & Dual-s-ST     & ——              & ——                        & ——                         & ——              & ——              & ——                        & ——                         & ——              \\
                            & Dual-s-JT     & \textbf{0.8570} & 0.8262                    & 0.9072                     & \textbf{0.0498} & 0.8286          & 0.7700                    & 0.8706                     & \textbf{0.0611} \\
                            & Single-s-JT   & 0.8564          & \textbf{0.8282}           & \textbf{0.9075}            & 0.0510          & \textbf{0.8293} & \textbf{0.7765}           & \textbf{0.8769}            & 0.0632          \\ \hline
\end{tabular}
        \label{tab:VSTSwinNetLSNet-RGB}} 
    \end{subtable}
    
    \begin{subtable}{\linewidth}
        \caption{RGB-D Datasets}
        \resizebox{\linewidth}{!}{
\begin{tabular}{cc|cccc|cccc}
\hline
\multicolumn{2}{c|}{\multirow{2}{*}{Model}} & \multicolumn{4}{c|}{NJUD}                                                                  & \multicolumn{4}{c}{SIP}                                                                    \\
\multicolumn{2}{c|}{}                       & $Sm\uparrow$    & $F_\beta ^{max} \uparrow$ & $E_\phi ^{\max } \uparrow$ & $MAE\downarrow$ & $Sm\uparrow$    & $F_\beta ^{max} \uparrow$ & $E_\phi ^{\max } \uparrow$ & $MAE\downarrow$ \\ \hline
\multirow{3}{*}{VST}        & Dual-s-ST     & 0.9224          & 0.9195                    & 0.9510                     & 0.0343          & 0.9036          & 0.9150                    & 0.9439                     & 0.0396          \\
                            & Dual-s-JT     & 0.9274          & 0.9241                    & 0.9540                     & 0.0324          & \textbf{0.9147} & \textbf{0.9231}           & \textbf{0.9516}            & \textbf{0.0342} \\
                            & Single-s-JT   & \textbf{0.9301} & \textbf{0.9310}           & \textbf{0.9589}            & \textbf{0.0314} & 0.9120          & 0.9222                    & 0.9504                     & 0.0354          \\ \hline
\multirow{3}{*}{SwinNet}    & Dual-s-ST     & 0.9255          & 0.9283                    & 0.9573                     & \textbf{0.0314} & 0.9009          & 0.9122                    & 0.9396                     & 0.0409          \\
                            & Dual-s-JT     & 0.9230          & 0.9277                    & 0.9537                     & 0.0338          & 0.9046          & 0.9158                    & 0.9440                     & 0.0385          \\
                            & Single-s-JT   & \textbf{0.9274} & \textbf{0.9305}           & \textbf{0.9578}            & 0.0316          & \textbf{0.9156} & \textbf{0.9318}           & \textbf{0.9537}            & \textbf{0.0332} \\ \hline
\multirow{3}{*}{LSNet}      & Dual-s-ST     & \textbf{0.9111} & \textbf{0.9144}           & \textbf{0.9498}            & \textbf{0.0386} & 0.8861          & \textbf{0.8952}           & \textbf{0.9306}            & \textbf{0.0496} \\
                            & Dual-s-JT     & 0.8786          & 0.8670                    & 0.9171                     & 0.0536          & 0.8730          & 0.8761                    & 0.9251                     & 0.0560          \\
                            & Single-s-JT   & 0.8907          & 0.8847                    & 0.9278                     & 0.0512          & \textbf{0.8864} & 0.8886                    & 0.9286                     & 0.0521          \\ \hline
\end{tabular}
        \label{tab:VSTSwinNetLSNet-RGBD}} 
    \end{subtable}

    \begin{subtable}{\linewidth}
        \caption{RGB-T Datasets}
        \resizebox{\linewidth}{!}{
\begin{tabular}{cc|cccc|cccc}
\hline
\multicolumn{2}{c|}{\multirow{2}{*}{Model}} & \multicolumn{4}{c|}{VT1000}                                                                & \multicolumn{4}{c}{VT5000}                                                                 \\
\multicolumn{2}{c|}{}                       & $Sm\uparrow$    & $F_\beta ^{max} \uparrow$ & $E_\phi ^{\max } \uparrow$ & $MAE\downarrow$ & $Sm\uparrow$    & $F_\beta ^{max} \uparrow$ & $E_\phi ^{\max } \uparrow$ & $MAE\downarrow$ \\ \hline
\multirow{3}{*}{VST}        & Dual-s-ST     & 0.9329          & 0.9314                    & 0.9705                     & 0.0211          & 0.8870          & 0.8610                    & 0.9286                     & 0.0383          \\
                            & Dual-s-JT     & \textbf{0.9418} & 0.9443                    & 0.9745                     & 0.0192          & \textbf{0.9048} & 0.8891                    & 0.9441                     & 0.0317          \\
                            & Single-s-JT   & \textbf{0.9418} & \textbf{0.9446}           & \textbf{0.9751}            & \textbf{0.0183} & \textbf{0.9048} & \textbf{0.8905}           & \textbf{0.9461}            & \textbf{0.0314} \\ \hline
\multirow{3}{*}{SwinNet}    & Dual-s-ST     & 0.9360          & 0.9392                    & 0.9727                     & 0.0179          & 0.9046          & 0.8920                    & 0.9481                     & 0.0290          \\
                            & Dual-s-JT     & 0.9365          & 0.9380                    & 0.9727                     & 0.0190          & 0.9119          & 0.9032                    & 0.9550                     & 0.0261          \\
                            & Single-s-JT   & \textbf{0.9449} & \textbf{0.9475}           & \textbf{0.9785}            & \textbf{0.0167} & \textbf{0.9191} & \textbf{0.9113}           & \textbf{0.9590}            & \textbf{0.0239} \\ \hline
\multirow{3}{*}{LSNet}      & Dual-s-ST     & 0.9256          & 0.9216                    & 0.9626                     & 0.0227          & 0.8774          & 0.8499                    & 0.9240                     & 0.0370          \\
                            & Dual-s-JT     & 0.9264          & 0.9222                    & 0.9613                     & 0.0229          & 0.8843          & 0.8614                    & 0.9288                     & \textbf{0.0364} \\
                            & Single-s-JT   & \textbf{0.9286} & \textbf{0.9266}           & \textbf{0.9644}            & \textbf{0.0224} & \textbf{0.8884} & \textbf{0.8708}           & \textbf{0.9338}            & 0.0367          \\ \hline
\end{tabular}
        \label{tab:VSTSwinNetLSNet-RGBT}} 
    \end{subtable}

\end{table}

We apply the proposed model framework to two transformer-based 3D SOD models, VST \cite{liu2021vst} and SwinNet \cite{liu2021swinnet}. 
In addition, a CNN-based LSNet \cite{zhou2023lsnet} (MobileV2Net \cite{MobileNetV2}) is used as a control for comparison.
VST, SwinNet, and LSNet are 3D SOD models, and VST also provides a version for RGB detection. 
Therefore, \cref{tab:VSTSwinNetLSNet} presents the results of VST, SwinNet, and LSNet on RGB-D and RGB-T datasets, as well as the results of VST on RGB datasets.
In \cref{tab:VSTSwinNetLSNet}, the first row of each model represents its original model architecture, and the first row of data is sourced from \cref{RGBRGBDRGBT}.

For the VST and SwinNet, the results of both models trained jointly (Dual-s-JT and Single-s-JT) outperform the results of the model trained with a single dataset (Dual-s-ST) on these datasets. Moreover, most metrics of the Single-s-JT model are superior to those of the Dual-s-JT model.
Since VST and SwinNet use the transformer network as the backbone, dual-stream VST and SwinNet with joint training (Dual-s-JT) still achieve relatively good results.
One of the learning networks of the dual-stream VST or SwinNet extracts RGB images, while the input to the other learning network is a mixture of RGB, depth, and thermal images. 
Although Dual-s-JT-VST and Dual-s-JT-SwinNet perform well on all three types of datasets, RGB saliency detection requires only a single-stream network. Therefore the proposed model framework is more suitable for saliency detection of RGB, RGB-D, and RGB-T data.  
It's worth noting that the single-stream VST model shows an approximate 18\% increase in training speed compared to the dual-stream VST model, and the model size decreases from 320MB to 238MB. Similarly, the single-stream SwinNet model exhibits about a 15\% increase in training speed compared to the dual-stream SwinNet model, with the model size decreasing from 786MB to 441MB. 
The proposed model framework has successfully migrated on the VST and SwinNet, providing a unified solution for RGB, RGB-D, and RGB-T SOD, resulting in improved model performance and reducing parameters.
Combining the experimental results of VST and SwinNet, the proposed model framework can be applied to other 3D SOD models, thus providing a unified solution for all three types of data and reducing model size.

Compared to the original LSNet (Dual-s-ST), LSNet (Single-s-JT)  using a single-stream network with joint training exhibits performance decrease on the RGB-D dataset and an improvement on the RGB-T dataset.
In \cref{tab:VSTSwinNetLSNet}, the LSNet trained jointly (Dual-s-JT and Single-s-JT ) outperforms the original LSNet (Dual-s-ST) in the RGB-T dataset. We infer that this result is likely due to the similarity between thermal and RGB images, as both are three-channel images with rich color information. Due to the large proportion of RGB data in the joint training dataset, the jointly trained model may perform better for RGB-T data because of this similarity. To confirm this inference, we conduct experiments by replacing the backbone network of AiOSOD with MobileV2Net \cite{MobileNetV2}. This model, adapts for MobileV2Net, removing the TFM and MFFM modules, adding  an additional FFM module. In \cref{tab8}, LSNet and AiOSOD both achieve optimal results by using a single-stream learning network and undergoing joint training. Moreover, when the training set contains only RGB-T data, Single-s-AiOSOD outperforms Dual-s-AiOSOD, and Single-s-LSNet performs close to Dual-s-LSNet. These experimental result confirms  our inference. Additionally, it can be observed that when (RGB, thermal) pairs are concatenated in the batch dimension, batch normalization has a minor impact on the prediction results, unlike the case with RGB-D.

\begin{table}[t]
\centering
\caption{LSNet and MobileV2Net-based AiOSOD detection results on RGB-T dataset. JT means joint training on RGB, RGB-D, and RGB-T datasets. Dual-s and Single-s indicate that the model employs a dual-stream backbone network  or a single-stream backbone network.}
\resizebox{\linewidth}{!}{
\begin{tabular}{cc|ccc|ccc}
\hline
\multicolumn{2}{c|}{\multirow{2}{*}{Model}}          & \multicolumn{3}{c|}{LSNet}          & \multicolumn{3}{c}{AiOSOD} \\
\multicolumn{2}{c|}{}                                & Dual-s & Single-s & Single-s        & Dual-s    & Single-s    & Single-s           \\
\multicolumn{2}{c|}{Train Dataset}                   & RGB-T  & RGB-T    & JT              & RGB-T     & RGB-T       & JT                 \\ \hline
\multirow{4}{*}{VT821}  & $Sm\uparrow$               & 0.8786 & 0.8783   & \textbf{0.9023} & 0.8408    & 0.8424      & \textbf{0.8984}    \\
                        & $F_\beta ^{max} \uparrow$  & 0.8448 & 0.8524   & \textbf{0.8872} & 0.7822    & 0.7813      & \textbf{0.8759}    \\
                        & $E_\phi ^{\max } \uparrow$ & 0.9205 & 0.9226   & \textbf{0.9431} & 0.8786    & 0.8749      & \textbf{0.9392}    \\
                        & $MAE\downarrow$            & 0.0332 & 0.0376   & \textbf{0.0290} & 0.0641    & 0.0570      & \textbf{0.0315}    \\ \hline
\multirow{4}{*}{VT1000} & $Sm\uparrow$               & 0.9256 & 0.9210   & \textbf{0.9286} & 0.9133    & 0.9167      & \textbf{0.9332}    \\
                        & $F_\beta ^{max} \uparrow$  & 0.9216 & 0.9186   & \textbf{0.9266} & 0.9070    & 0.9094      & \textbf{0.9329}    \\
                        & $E_\phi ^{\max } \uparrow$ & 0.9626 & 0.9618   & \textbf{0.9644} & 0.9535    & 0.9552      & \textbf{0.9704}    \\
                        & $MAE\downarrow$            & 0.0227 & 0.0262   & \textbf{0.0224} & 0.0289    & 0.0288      & \textbf{0.0234}    \\ \hline
\multirow{4}{*}{VT5000} & $Sm\uparrow$               & 0.8774 & 0.8784   & \textbf{0.8884} & 0.8496    & 0.8496      & \textbf{0.8883}    \\
                        & $F_\beta ^{max} \uparrow$  & 0.8499 & 0.8553   & \textbf{0.8708} & 0.8035    & 0.8043      & \textbf{0.8647}    \\
                        & $E_\phi ^{\max } \uparrow$ & 0.9240 & 0.9313   & \textbf{0.9338} & 0.8955    & 0.8957      & \textbf{0.9337}    \\
                        & $MAE\downarrow$            & 0.0370 & 0.0387   & \textbf{0.0367} & 0.0511    & 0.0513      & \textbf{0.0380}    \\ \hline
\end{tabular}
}
\label{tab8}
\end{table}

\section{Conclusions}
In this paper, we are the first  to consider a unified solution to realize RGB, RGB-D and RGB-T SOD, requiring only one model and  same weights to perform SOD on all three modalities. To achieve the unified solution, we propose a model framework and develop a lightweight model for validation (AiOSOD). The lightweight AiOSOD model demonstrates excellent  performance on RGB, RGB-D, and RGB-T datasets, effectively balancing performance and speed. Our proposed model framework takes three types of data as the training set,  concatenating them in the batch dimension and extracting features through a transformer network. With this framework, the model can learn from all three types of data and avoid performance degradation due to interference between multimodal features. Importantly, the proposed model framework can be applied to other 3D SOD models, reducing model size and providing a unified solution for RGB, RGB-D, and RGB-T SOD. In the future, we will continue to explore more efficient unified solutions for SOD.


\begin{thebibliography}{45}
\providecommand{\natexlab}[1]{#1}
\providecommand{\url}[1]{\texttt{#1}}
\expandafter\ifx\csname urlstyle\endcsname\relax
  \providecommand{\doi}[1]{doi: #1}\else
  \providecommand{\doi}{doi: \begingroup \urlstyle{rm}\Url}\fi

\bibitem[Borji et~al.(2015)Borji, Cheng, Jiang, and Li]{mae-fm}
Ali Borji, Ming-Ming Cheng, Huaizu Jiang, and Jia Li.
\newblock Salient object detection: A benchmark.
\newblock \emph{IEEE Transactions on Image Processing}, 24\penalty0
  (12):\penalty0 5706--5722, 2015.

\bibitem[Cheng et~al.(2022)Cheng, Zheng, Pei, Tang, Lyu, and Chen]{DIGR-Net}
Xiaolong Cheng, Xuan Zheng, Jialun Pei, He Tang, Zehua Lyu, and Chuanbo Chen.
\newblock Depth-induced gap-reducing network for rgb-d salient object
  detection: An interaction, guidance and refinement approach.
\newblock \emph{IEEE Transactions on Multimedia}, pages 1--1, 2022.

\bibitem[Cong et~al.(2022)Cong, Zhang, Zhang, Zheng, Zhao, Huang, and
  Kwong]{tnet}
Runmin Cong, Kepu Zhang, Chen Zhang, Feng Zheng, Yao Zhao, Qingming Huang, and
  Sam Kwong.
\newblock Does thermal really always matter for rgb-t salient object detection?
\newblock \emph{IEEE Transactions on Multimedia}, pages 1--12, 2022.

\bibitem[Dosovitskiy et~al.(2020)Dosovitskiy, Beyer, Kolesnikov, Weissenborn,
  Zhai, Unterthiner, Dehghani, Minderer, Heigold, Gelly,
  et~al.]{dosovitskiy2020vit}
Alexey Dosovitskiy, Lucas Beyer, Alexander Kolesnikov, Dirk Weissenborn,
  Xiaohua Zhai, Thomas Unterthiner, Mostafa Dehghani, Matthias Minderer, Georg
  Heigold, Sylvain Gelly, et~al.
\newblock An image is worth 16x16 words: Transformers for image recognition at
  scale.
\newblock \emph{arXiv preprint arXiv:2010.11929}, 2020.

\bibitem[Fan et~al.(2017)Fan, Cheng, Liu, Li, and Borji]{Smeasure}
Deng-Ping Fan, Ming-Ming Cheng, Yun Liu, Tao Li, and Ali Borji.
\newblock Structure-measure: A new way to evaluate foreground maps.
\newblock In \emph{Proceedings of the IEEE International Conference on Computer
  Vision (ICCV)}, 2017.

\bibitem[Fan et~al.(2018)Fan, Gong, Cao, Ren, Cheng, and Borji]{Emeasure}
Deng-Ping Fan, Cheng Gong, Yang Cao, Bo Ren, Ming-Ming Cheng, and Ali Borji.
\newblock Enhanced-alignment measure for binary foreground map evaluation.
\newblock In \emph{Proceedings of the Twenty-Seventh International Joint
  Conference on Artificial Intelligence, {IJCAI-18}}, pages 698--704.
  International Joint Conferences on Artificial Intelligence Organization,
  2018.

\bibitem[Fan et~al.(2019)Fan, Wang, Cheng, and Shen]{compression}
Deng-Ping Fan, Wenguan Wang, Ming-Ming Cheng, and Jianbing Shen.
\newblock Shifting more attention to video salient object detection.
\newblock In \emph{Proceedings of the IEEE/CVF Conference on Computer Vision
  and Pattern Recognition (CVPR)}, 2019.

\bibitem[Fan et~al.(2021)Fan, Lin, Zhang, Zhu, and Cheng]{SIP}
Deng-Ping Fan, Zheng Lin, Zhao Zhang, Menglong Zhu, and Ming-Ming Cheng.
\newblock Rethinking rgb-d salient object detection: Models, data sets, and
  large-scale benchmarks.
\newblock \emph{IEEE Transactions on Neural Networks and Learning Systems},
  32\penalty0 (5):\penalty0 2075--2089, 2021.

\bibitem[Fu et~al.(2022)Fu, Fan, Ji, Zhao, Shen, and Zhu]{fu2021jldcf}
Keren Fu, Deng-Ping Fan, Ge-Peng Ji, Qijun Zhao, Jianbing Shen, and Ce Zhu.
\newblock Siamese network for rgb-d salient object detection and beyond.
\newblock \emph{IEEE Transactions on Pattern Analysis and Machine
  Intelligence}, 44\penalty0 (9):\penalty0 5541--5559, 2022.

\bibitem[Gao et~al.(2015)Gao, Shi, Tao, and Xu]{retrieval}
Yuan Gao, Miaojing Shi, Dacheng Tao, and Chao Xu.
\newblock Database saliency for fast image retrieval.
\newblock \emph{IEEE Transactions on Multimedia}, 17\penalty0 (3):\penalty0
  359--369, 2015.

\bibitem[Guo and Zhang(2010)]{segmentation2}
Chenlei Guo and Liming Zhang.
\newblock A novel multiresolution spatiotemporal saliency detection model and
  its applications in image and video compression.
\newblock \emph{IEEE Transactions on Image Processing}, 19\penalty0
  (1):\penalty0 185--198, 2010.

\bibitem[Huo et~al.(2022)Huo, Zhu, Zhang, Liu, and Shu]{CSRNet}
Fushuo Huo, Xuegui Zhu, Lei Zhang, Qifeng Liu, and Yu Shu.
\newblock Efficient context-guided stacked refinement network for rgb-t salient
  object detection.
\newblock \emph{IEEE Transactions on Circuits and Systems for Video
  Technology}, 32\penalty0 (5):\penalty0 3111--3124, 2022.

\bibitem[Jia et~al.(2022)Jia, DongYe, and Peng]{jia2022siatrans}
XingZhao Jia, ChangLei DongYe, and YanJun Peng.
\newblock Siatrans: Siamese transformer network for rgb-d salient object
  detection with depth image classification.
\newblock \emph{Image and Vision Computing}, 127:\penalty0 104549, 2022.

\bibitem[Ju et~al.(2014)Ju, Ge, Geng, Ren, and Wu]{njud}
Ran Ju, Ling Ge, Wenjing Geng, Tongwei Ren, and Gangshan Wu.
\newblock Depth saliency based on anisotropic center-surround difference.
\newblock In \emph{2014 IEEE International Conference on Image Processing
  (ICIP)}, pages 1115--1119, 2014.

\bibitem[Ke and Tsubono(2022)]{RCSB}
Yun~Yi Ke and Takahiro Tsubono.
\newblock Recursive contour-saliency blending network for accurate salient
  object detection.
\newblock In \emph{Proceedings of the IEEE/CVF Winter Conference on
  Applications of Computer Vision (WACV)}, pages 2940--2950, 2022.

\bibitem[Kingma and Ba(2014)]{kingma2014adam}
Diederik~P Kingma and Jimmy Ba.
\newblock Adam: A method for stochastic optimization.
\newblock \emph{arXiv preprint arXiv:1412.6980}, 2014.

\bibitem[Liao et~al.(2022)Liao, Gao, Li, Wang, and Kwong]{liao2022ccfenet}
Guibiao Liao, Wei Gao, Ge Li, Junle Wang, and Sam Kwong.
\newblock Cross-collaborative fusion-encoder network for robust rgb-thermal
  salient object detection.
\newblock \emph{IEEE Transactions on Circuits and Systems for Video
  Technology}, 32\penalty0 (11):\penalty0 7646--7661, 2022.

\bibitem[Liu et~al.(2021{\natexlab{a}})Liu, Liu, Peng, and Cheng]{CII}
Jiang-Jiang Liu, Zhi-Ang Liu, Pai Peng, and Ming-Ming Cheng.
\newblock Rethinking the u-shape structure for salient object detection.
\newblock \emph{IEEE Transactions on Image Processing}, 30:\penalty0
  9030--9042, 2021{\natexlab{a}}.

\bibitem[Liu et~al.(2021{\natexlab{b}})Liu, Zhang, Wan, Shao, and
  Han]{liu2021vst}
Nian Liu, Ni Zhang, Kaiyuan Wan, Ling Shao, and Junwei Han.
\newblock Visual saliency transformer.
\newblock In \emph{Proceedings of the IEEE/CVF International Conference on
  Computer Vision (ICCV)}, pages 4722--4732, 2021{\natexlab{b}}.

\bibitem[Liu et~al.(2021{\natexlab{c}})Liu, Lin, Cao, Hu, Wei, Zhang, Lin, and
  Guo]{Swintransformer}
Ze Liu, Yutong Lin, Yue Cao, Han Hu, Yixuan Wei, Zheng Zhang, Stephen Lin, and
  Baining Guo.
\newblock Swin transformer: Hierarchical vision transformer using shifted
  windows.
\newblock In \emph{Proceedings of the IEEE/CVF International Conference on
  Computer Vision (ICCV)}, pages 10012--10022, 2021{\natexlab{c}}.

\bibitem[Liu et~al.(2022)Liu, Tan, He, and Xiao]{liu2021swinnet}
Zhengyi Liu, Yacheng Tan, Qian He, and Yun Xiao.
\newblock Swinnet: Swin transformer drives edge-aware rgb-d and rgb-t salient
  object detection.
\newblock \emph{IEEE Transactions on Circuits and Systems for Video
  Technology}, 32\penalty0 (7):\penalty0 4486--4497, 2022.

\bibitem[Pang et~al.(2023)Pang, Zhao, Zhang, and Lu]{CAVER}
Youwei Pang, Xiaoqi Zhao, Lihe Zhang, and Huchuan Lu.
\newblock Caver: Cross-modal view-mixed transformer for bi-modal salient object
  detection.
\newblock \emph{IEEE Transactions on Image Processing}, 32:\penalty0 892--904,
  2023.

\bibitem[Paszke et~al.(2019)Paszke, Gross, Massa, Lerer, Bradbury, Chanan,
  Killeen, Lin, Gimelshein, Antiga, et~al.]{paszke2019pytorch}
Adam Paszke, Sam Gross, Francisco Massa, Adam Lerer, James Bradbury, Gregory
  Chanan, Trevor Killeen, Zeming Lin, Natalia Gimelshein, Luca Antiga, et~al.
\newblock Pytorch: An imperative style, high-performance deep learning library.
\newblock \emph{Advances in neural information processing systems}, 32, 2019.

\bibitem[Peng et~al.(2014)Peng, Li, Xiong, Hu, and Ji]{NLPR}
Houwen Peng, Bing Li, Weihua Xiong, Weiming Hu, and Rongrong Ji.
\newblock Rgbd salient object detection: A benchmark and algorithms.
\newblock In \emph{Computer Vision -- ECCV 2014}, pages 92--109, Cham, 2014.
  Springer International Publishing.

\bibitem[Piao et~al.(2019)Piao, Ji, Li, Zhang, and Lu]{DUTLF-D}
Yongri Piao, Wei Ji, Jingjing Li, Miao Zhang, and Huchuan Lu.
\newblock Depth-induced multi-scale recurrent attention network for saliency
  detection.
\newblock In \emph{Proceedings of the IEEE/CVF International Conference on
  Computer Vision (ICCV)}, 2019.

\bibitem[Sandler et~al.(2018)Sandler, Howard, Zhu, Zhmoginov, and
  Chen]{MobileNetV2}
Mark Sandler, Andrew Howard, Menglong Zhu, Andrey Zhmoginov, and Liang-Chieh
  Chen.
\newblock Mobilenetv2: Inverted residuals and linear bottlenecks.
\newblock In \emph{Proceedings of the IEEE Conference on Computer Vision and
  Pattern Recognition (CVPR)}, 2018.

\bibitem[Tu et~al.(2020)Tu, Xia, Li, Wang, Ma, and Tang]{VT1000}
Zhengzheng Tu, Tian Xia, Chenglong Li, Xiaoxiao Wang, Yan Ma, and Jin Tang.
\newblock Rgb-t image saliency detection via collaborative graph learning.
\newblock \emph{IEEE Transactions on Multimedia}, 22\penalty0 (1):\penalty0
  160--173, 2020.

\bibitem[Tu et~al.(2022)Tu, Ma, Li, Li, Xu, and Liu]{VT5000}
Zhengzheng Tu, Yan Ma, Zhun Li, Chenglong Li, Jieming Xu, and Yongtao Liu.
\newblock Rgbt salient object detection: A large-scale dataset and benchmark.
\newblock \emph{IEEE Transactions on Multimedia}, pages 1--1, 2022.

\bibitem[Vaswani et~al.(2017)Vaswani, Shazeer, Parmar, Uszkoreit, Jones, Gomez,
  Kaiser, and Polosukhin]{vaswani2017attention}
Ashish Vaswani, Noam Shazeer, Niki Parmar, Jakob Uszkoreit, Llion Jones,
  Aidan~N Gomez, {\L}ukasz Kaiser, and Illia Polosukhin.
\newblock Attention is all you need.
\newblock \emph{Advances in neural information processing systems}, 30, 2017.

\bibitem[Wang et~al.(2018)Wang, Li, Ma, Zheng, Tang, and Luo]{VT821}
Guizhao Wang, Chenglong Li, Yunpeng Ma, Aihua Zheng, Jin Tang, and Bin Luo.
\newblock Rgb-t saliency detection benchmark: Dataset, baselines, analysis and
  a novel approach.
\newblock In \emph{Image and Graphics Technologies and Applications}, pages
  359--369, Singapore, 2018. Springer Singapore.

\bibitem[Wang et~al.(2017)Wang, Lu, Wang, Feng, Wang, Yin, and Ruan]{DUTS}
Lijun Wang, Huchuan Lu, Yifan Wang, Mengyang Feng, Dong Wang, Baocai Yin, and
  Xiang Ruan.
\newblock Learning to detect salient objects with image-level supervision.
\newblock In \emph{Proceedings of the IEEE Conference on Computer Vision and
  Pattern Recognition (CVPR)}, 2017.

\bibitem[Wang et~al.(2022)Wang, Xie, Li, Fan, Song, Liang, Lu, Luo, and
  Shao]{wang2022pvt}
Wenhai Wang, Enze Xie, Xiang Li, Deng-Ping Fan, Kaitao Song, Ding Liang, Tong
  Lu, Ping Luo, and Ling Shao.
\newblock Pvt v2: Improved baselines with pyramid vision transformer.
\newblock \emph{Computational Visual Media}, 8\penalty0 (3):\penalty0 415--424,
  2022.

\bibitem[Woo et~al.(2018)Woo, Park, Lee, and Kweon]{woo2018cbam}
Sanghyun Woo, Jongchan Park, Joon-Young Lee, and In~So Kweon.
\newblock Cbam: Convolutional block attention module.
\newblock In \emph{Proceedings of the European conference on computer vision
  (ECCV)}, pages 3--19, 2018.

\bibitem[Wu et~al.(2022)Wu, Liu, Xu, Bian, Gu, and Cheng]{MobileSal}
Yu-Huan Wu, Yun Liu, Jun Xu, Jia-Wang Bian, Yu-Chao Gu, and Ming-Ming Cheng.
\newblock Mobilesal: Extremely efficient rgb-d salient object detection.
\newblock \emph{IEEE Transactions on Pattern Analysis and Machine
  Intelligence}, 44\penalty0 (12):\penalty0 10261--10269, 2022.

\bibitem[Yan et~al.(2013)Yan, Xu, Shi, and Jia]{ECSSD}
Qiong Yan, Li Xu, Jianping Shi, and Jiaya Jia.
\newblock Hierarchical saliency detection.
\newblock In \emph{Proceedings of the IEEE Conference on Computer Vision and
  Pattern Recognition (CVPR)}, 2013.

\bibitem[Yang et~al.(2013)Yang, Zhang, Lu, Ruan, and Yang]{DUT-OMORN}
Chuan Yang, Lihe Zhang, Huchuan Lu, Xiang Ruan, and Ming-Hsuan Yang.
\newblock Saliency detection via graph-based manifold ranking.
\newblock In \emph{Proceedings of the IEEE Conference on Computer Vision and
  Pattern Recognition (CVPR)}, 2013.

\bibitem[Ye et~al.(2017)Ye, Liu, Li, Shen, Bai, and Wang]{segmentation1}
Linwei Ye, Zhi Liu, Lina Li, Liquan Shen, Cong Bai, and Yang Wang.
\newblock Salient object segmentation via effective integration of saliency and
  objectness.
\newblock \emph{IEEE Transactions on Multimedia}, 19\penalty0 (8):\penalty0
  1742--1756, 2017.

\bibitem[Yuan et~al.(2021)Yuan, Chen, Wang, Yu, Shi, Jiang, Tay, Feng, and
  Yan]{yuan2021t2tvit}
Li Yuan, Yunpeng Chen, Tao Wang, Weihao Yu, Yujun Shi, Zi-Hang Jiang,
  Francis~E.H. Tay, Jiashi Feng, and Shuicheng Yan.
\newblock Tokens-to-token vit: Training vision transformers from scratch on
  imagenet.
\newblock In \emph{Proceedings of the IEEE/CVF International Conference on
  Computer Vision (ICCV)}, pages 558--567, 2021.

\bibitem[Zhang et~al.(2022)Zhang, Yao, Hu, Piao, and Ji]{C2DFNet}
Miao Zhang, Shunyu Yao, Beiqi Hu, Yongri Piao, and Wei Ji.
\newblock C$^{2}$dfnet: Criss-cross dynamic filter network for rgb-d salient
  object detection.
\newblock \emph{IEEE Transactions on Multimedia}, pages 1--13, 2022.

\bibitem[Zhang et~al.(2021)Zhang, Ji, Wang, Fu, and Zhao]{DFM-Net}
Wenbo Zhang, Ge-Peng Ji, Zhuo Wang, Keren Fu, and Qijun Zhao.
\newblock Depth quality-inspired feature manipulation for efficient rgb-d
  salient object detection.
\newblock In \emph{Proceedings of the 29th ACM International Conference on
  Multimedia}, page 731–740, New York, NY, USA, 2021. Association for
  Computing Machinery.

\bibitem[Zhang et~al.(2023)Zhang, Wang, Yang, Zhang, Gong, and
  Wang]{zhang2023csnet}
Yunhua Zhang, Hangxu Wang, Gang Yang, Jianhao Zhang, Congjin Gong, and Yutao
  Wang.
\newblock Csnet: a convnext-based siamese network for rgb-d salient object
  detection.
\newblock \emph{The Visual Computer}, pages 1--19, 2023.

\bibitem[Zhao et~al.(2021)Zhao, Xia, Xie, and Li]{CTDNet}
Zhirui Zhao, Changqun Xia, Chenxi Xie, and Jia Li.
\newblock Complementary trilateral decoder for fast and accurate salient object
  detection.
\newblock In \emph{Proceedings of the 29th ACM International Conference on
  Multimedia}, page 4967–4975, New York, NY, USA, 2021. Association for
  Computing Machinery.

\bibitem[Zhou et~al.(2020)Zhou, Xie, Lai, Chen, and Yang]{ITSD}
Huajun Zhou, Xiaohua Xie, Jian-Huang Lai, Zixuan Chen, and Lingxiao Yang.
\newblock Interactive two-stream decoder for accurate and fast saliency
  detection.
\newblock In \emph{Proceedings of the IEEE/CVF Conference on Computer Vision
  and Pattern Recognition (CVPR)}, 2020.

\bibitem[Zhou et~al.(2023)Zhou, Zhu, Lei, Yang, and Yu]{zhou2023lsnet}
Wujie Zhou, Yun Zhu, Jingsheng Lei, Rongwang Yang, and Lu Yu.
\newblock Lsnet: Lightweight spatial boosting network for detecting salient
  objects in rgb-thermal images.
\newblock \emph{IEEE Transactions on Image Processing}, 32:\penalty0
  1329--1340, 2023.

\bibitem[Zhuge et~al.(2023)Zhuge, Fan, Liu, Zhang, Xu, and Shao]{ICON}
Mingchen Zhuge, Deng-Ping Fan, Nian Liu, Dingwen Zhang, Dong Xu, and Ling Shao.
\newblock Salient object detection via integrity learning.
\newblock \emph{IEEE Transactions on Pattern Analysis and Machine
  Intelligence}, 45\penalty0 (3):\penalty0 3738--3752, 2023.

\end{thebibliography}
\end{document}